\documentclass[sn-apa, maxcitenames=1,mincitenames=1]{sn-jnl}

\usepackage{enumitem}
\usepackage{setspace}

\usepackage[table]{xcolor}
\usepackage[utf8]{inputenc}
\usepackage[english]{babel}
\usepackage{algorithm}
\usepackage{algpseudocode}

\usepackage{amsmath}
\usepackage{graphicx}
\usepackage{amssymb}
\usepackage[table]{xcolor}

\usepackage{caption}
\usepackage{subcaption}
\usepackage{textcomp}
\usepackage{algorithm}
\usepackage{algpseudocode}
\usepackage{graphicx}%
\usepackage{multirow}%
\usepackage{amsmath,amssymb,amsfonts}%
\usepackage{amsthm}%
\usepackage{mathrsfs}%
\usepackage[title]{appendix}%
\usepackage{xcolor}%
\usepackage{textcomp}%
\usepackage{manyfoot}%
\usepackage{booktabs}%
\usepackage{algorithm}%
\usepackage{algorithmicx}%
\usepackage{algpseudocode}%
\usepackage{listings}%
\usepackage{geometry}
\theoremstyle{thmstyleone}%
\setcitestyle{authoryear, maxnames=1}
\theoremstyle{thmstyletwo}%
\newtheorem{remark}{Remark}%

\theoremstyle{thmstylethree}%
\begin{document}
\onehalfspacing  

\title[Title]{A Learning-Based Superposition Operator for Non-Renewal Arrival Processes in Queueing Networks}


\author*[1]{\fnm{Eliran} \sur{Sherzer}}\email{eliransh@ariel.ac.il}

\affil*[1]{\orgdiv{Industrial Engineering}, \orgname{Ariel University}, \orgaddress{\street{Ramat HaGolan St 65}, \city{Ariel}, \postcode{4070000}, \country{Israel}}}


\abstract{ The superposition of arrival processes is a fundamental yet analytically intractable operation in queueing networks when inputs are general non-renewal streams. Classical methods either reduce merged flows to renewal surrogates, rely on computationally prohibitive Markovian representations, or focus solely on mean-value performance measures. 

We propose a scalable data-driven superposition operator that maps low-order moments and autocorrelation descriptors of multiple arrival streams to those of their merged process. The operator is a deep learning model trained on synthetically generated Markovian Arrival Processes (MAPs), for which exact superposition is available, and learns a compact representation that accurately reconstructs the first five moments and short-range dependence structure of the aggregate stream. Extensive computational experiments demonstrate uniformly low prediction errors across heterogeneous variability and correlation regimes, substantially outperforming classical renewal-based approximations.

When integrated with learning-based modules for departure-process and steady-state analysis, the proposed operator enables decomposition-based evaluation of feed-forward queueing networks with merging flows. The framework provides a scalable alternative to traditional analytical approaches while preserving higher-order variability and dependence information required for accurate distributional performance analysis.

}




\maketitle

\section{Introduction}\label{sec:intro}

The superposition of arrival processes is a fundamental operation in queueing systems and stochastic networks.
In many practical systems, multiple independent traffic streams merge at service nodes, communication links, or processing centers, forming aggregate arrival processes that govern congestion, delay, and stability.
Such merging phenomena naturally arise in open queueing networks, manufacturing systems, communication networks, and service systems with multiple upstream sources.
Accurate characterization of superposed arrival processes is therefore essential for meaningful queueing analysis \cite{Whitt1982,Newell1984}.

Despite its importance, the superposition of general arrival processes remains analytically intractable in most cases.
While the superposition of Poisson processes remains Poisson, and the superposition of Markovian Arrival Processes (MAPs) admits an exact representation through Kronecker constructions \cite{Neuts1979,HeNeuts1998}, the superposition of renewal or general non-renewal processes is neither renewal nor Markovian in general.
As a consequence, closed-form expressions for inter-arrival time distributions, higher-order moments, or temporal correlations of merged streams are typically unavailable.

This lack of structural closure creates a fundamental limitation for operational performance analysis.
Many performance measures, such as steady-state queue-length distributions, waiting-time tails, and departure-process statistics, depend not only on mean arrival rates but also on higher-order moments and temporal correlation structure \cite{GlynnWhitt1994,Asmussen2003}.
When superposition discards such information, downstream performance predictions may become unreliable, particularly in heavily loaded systems and multi-node networks where dependence propagates across stations.

Classical superposition approximations, including those of~\cite{Whitt1982} and~\cite{Albin1984}, provide practical rules for merging renewal and mildly dependent streams.
However, these methods typically summarize the aggregate process using only low-dimensional variability descriptors, most notably the mean rate and squared coefficient of variation (SCV).
While suitable for first-order performance estimates, such reductions neglect higher-order variability and serial dependence, limiting their ability to support distributional queueing analysis.

A related line of work relies on variability indices, such as the index of dispersion for counts (IDC), as developed in the robust queueing network analyzer of~\cite{https://doi.org/10.1002/nav.22010}.
These approaches propagate time-scale-dependent variability measures through network nodes and enable computationally efficient mean-performance approximations.
Nevertheless, they remain primarily limited to first-order metrics (e.g., mean queue length or delay) and do not yield full stationary distributions.

Diffusion and heavy-traffic approximations offer another perspective \cite{Reiman1984}, replacing superposed non-renewal inputs with Gaussian processes under appropriate scaling.
Although mathematically elegant, these models preserve only first- and second-order variability information and typically approximate workload or mean queue-length behavior rather than full steady-state customer-count distributions across general operating regimes.

Taken together, existing approaches either 
(i) compress the merged stream into low-order variability summaries, 
(ii) rely on exact but computationally prohibitive Markovian representations, or 
(iii) focus exclusively on mean-value performance measures.
A scalable framework that preserves higher-order variability and dependence information while enabling distributional queueing analysis remains absent from the literature.

In this paper, we introduce a \emph{learning-based superposition operator} for non-renewal arrival processes via  a neural network (NN).
The key contribution is not a black-box performance predictor, but a structural mapping that restores tractability within decomposition-based queueing analysis.
Rather than attempting to represent arrival streams via explicit stochastic process models, we directly approximate the \emph{moments and autocorrelation structure} of the merged process.
Specifically, given the first $n$ moments and selected autocorrelation coefficients of each input stream, we approximate the first five moments and short-range dependence descriptors of their superposition.

At first glance, approximating only low-order statistical descriptors may appear restrictive.
However, as demonstrated below and building on the NN based framework of~\cite{SHERZER2025141}, these descriptors are sufficient to support accurate steady-state and departure-process approximation at downstream stations.
Empirically, we show that the superposition mapping is identifiable from a compact representation involving moderate-order moments and short-range dependence.
This compact operator perspective enables scalable queueing network analysis without requiring parametric process fitting or explicit state-space construction.

The core idea is to learn this operator from exact data.
We generate a large and diverse family of MAPs whose variability and dependence structures span wide regimes.
Using the exact Kronecker-based superposition of~\cite{Neuts1979}, we compute the merged MAP corresponding to each pair of input streams and extract its statistical descriptors.
These input-output descriptor pairs serve as labeled data for training a neural network that learns the superposition mapping.

MAPs are used solely as a data-generation mechanism.
Because MAPs are dense in the class of nonnegative distributions and can approximate a broad family of stationary point processes \cite{Asmussen2003,HorvathTelek2002}, they provide a flexible domain over which to learn the operator.
Importantly, the trained superposition module does \emph{not} require an explicit MAP representation at inference time.
At deployment, the operator acts directly on moment and autocorrelation descriptors of arbitrary arrival processes.
Thus, the framework generalizes beyond MAP inputs and avoids the state-space explosion inherent in exact MAP superposition.

Although the present work focuses on superposition as a standalone component, this operator serves as a foundational building block for open queueing network analysis.
When combined with previously developed neural modules for departure-process characterization and steady-state distribution approximation~\cite{SHERZER2025141}, the superposition operator enables recursive decomposition of feed-forward queueing networks with merging flows.
This extension removes the structural restriction to tandem systems and makes it possible, for the first time, to approximate full steady-state distributions in non-Markovian networks with merging streams in a scalable manner.

Crucially, recursive superposition allows aggregation of any finite number of arrival streams, thereby supporting general feed-forward single-server architectures in which multiple correlated streams may enter each node.

\begin{figure}
\centering
\includegraphics[scale = 0.65]{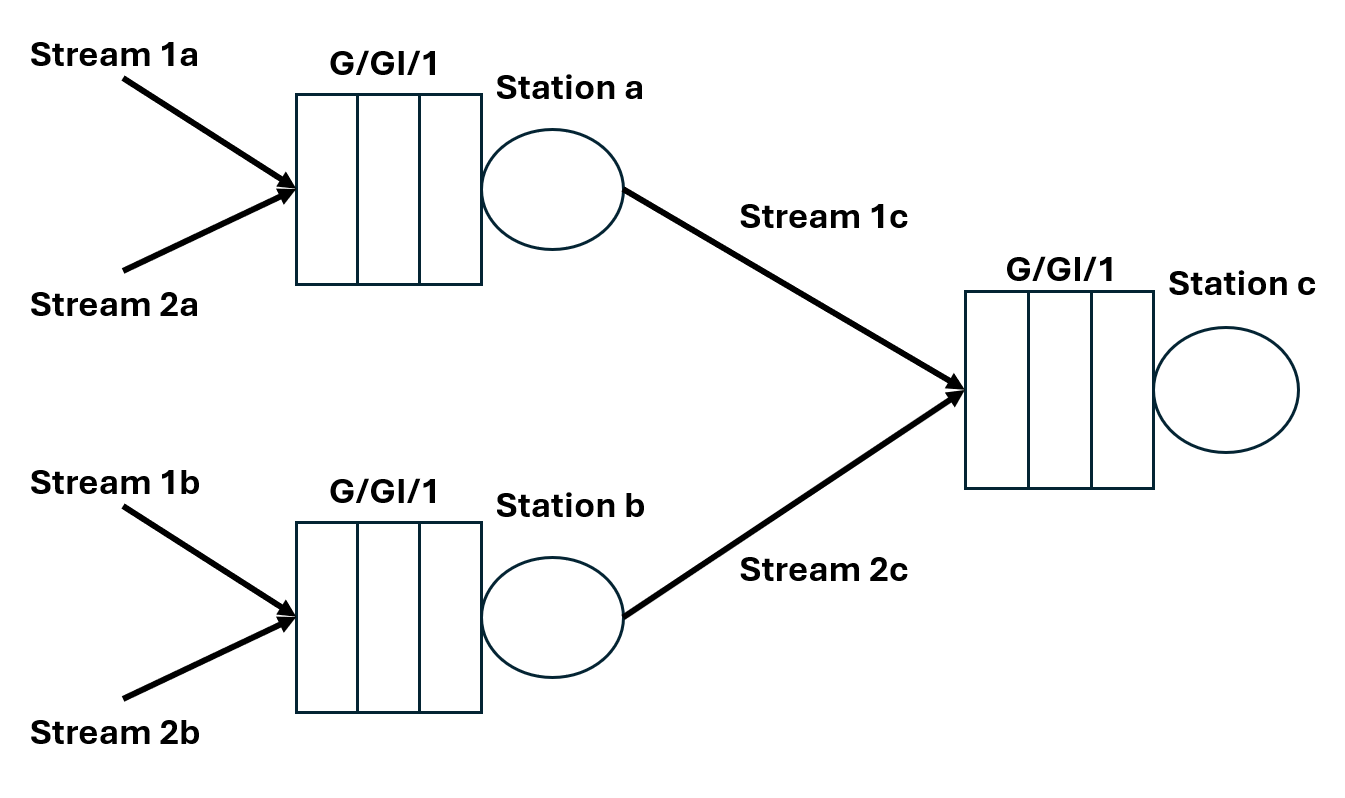}
\caption{Queueing network topology.}
\label{fig:system2}
\end{figure}

For illustration, we consider the three-station topology depicted in Figure~\ref{fig:system2}. 
Stations $a$ and $b$ each receive two external, possibly non-renewal arrival streams, while service times at all stations follow known renewal processes. 
Departures from Stations $a$ and $b$ are routed to Station $c$, where our objective is to approximate the steady-state distribution.

\begin{figure}
\centering
\includegraphics[scale = 0.55]{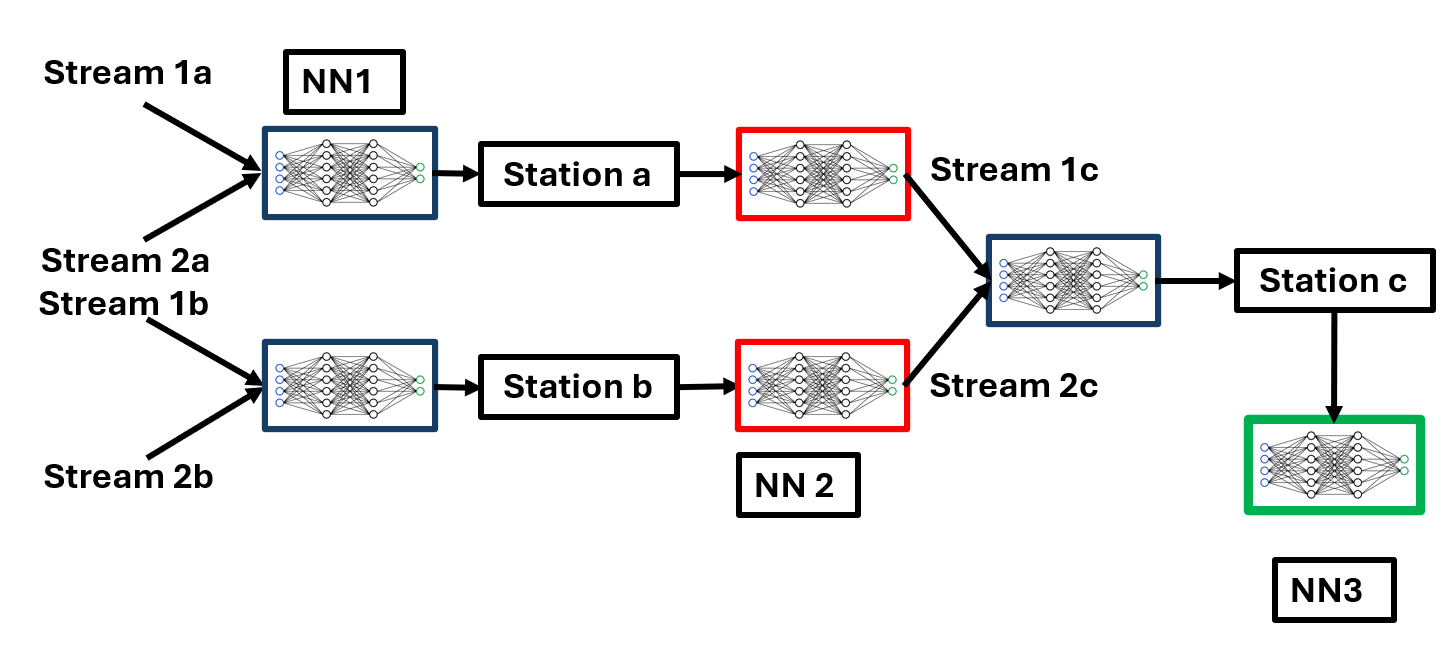}
\caption{Queueing network via NNs.  }
\label{fig:tier2_nns}
\end{figure}

Under the decomposition framework, the analysis proceeds sequentially and is governed by three neural modules, illustrated in Figure~\ref{fig:tier2_nns}. 
First, the external streams entering Stations $a$ and $b$ are merged using  \textbf{NN 1} (blue rectangle), which implements the neural superposition operator. 
Next, the non-renewal departure processes from Stations $a$ and $b$ are approximated via  \textbf{NN 2} (red rectangle), producing the required moment and autocorrelation descriptors. 
The resulting departure streams are then superposed again using  \textbf{NN 1}. 
Finally, the superposed arrival process, together with the service-time descriptors at Station $c$, is provided to  \textbf{NN 3} (green rectangle), which approximates the steady-state distribution of the number of customers at Station $c$. For clarification only  \textbf{NN 1} is developed in this paper, while   \textbf{NN 2} and   \textbf{NN 3} were develpoed in~\cite{SHERZER2025141}.

To summarize, the proposed framework consists of three modular components:
(i) a neural superposition operator,
(ii) a departure-process approximation module, and
(iii) a steady-state distribution predictor.
Together, these modules enable scalable distributional analysis of feed-forward queueing networks with merging flows.
To our knowledge, no existing framework simultaneously (a) merges general renewal and non-renewal streams, (b) preserves higher-order variability and dependence in compact form, and (c) enables downstream steady-state distribution approximation without parametric modeling or state-space explosion.

The main contributions of this paper are threefold.
First, we introduce a neural-network-based superposition operator that approximates higher-order moments and dependence descriptors of merged non-renewal streams.
Second, we provide an open-source implementation (\url{https://github.com/eliransher/superposition-of-non-renewal-procesess.git}) enabling fast inference.
Third, we demonstrate that this operator extends the class of tractable non-Markovian queueing networks beyond tandem structures, thereby restoring decomposition-based analysis in settings previously considered analytically inaccessible.

The remainder of the paper is organized as follows.
Section~\ref{sec:lit} reviews the relevant literature.
Section~\ref{sec:prob_form} formalizes the superposition operator and descriptor framework.
Section~\ref{sec:training_proc} outlines the data-generation and training procedures.
Sections~\ref{sec:experiments}-\ref{sec:results} present the experimental design and numerical results.
Section~\ref{sec:questa} demonstrates the integration of the operator within a decomposition-based queueing network analysis.
Section~\ref{sec:conclu} concludes and discusses limitations and future research directions.

\section{Literature Review}
\label{sec:lit}

As discussed in the Introduction, the superposition of arrival processes plays a fundamental role in queueing networks but remains analytically challenging outside special cases. While Poisson inputs are closed under superposition and MAPs admit an exact Kronecker-based construction \cite{Neuts1979,HeNeuts1998}, no comparable structural closure exists for renewal or general non-renewal processes. The resulting merged stream is typically neither renewal nor Markovian, and key statistical descriptors-particularly higher-order moments and temporal correlations-are not available in closed form. This limitation is critical, as such descriptors are known to strongly affect congestion and delay behavior in queueing systems \cite{Asmussen2003,GlynnWhitt1994}.

A classical approach to merging non-renewal arrival streams is to replace each
input process by a renewal surrogate, and then analyze the superposition of the
surrogates.
\cite{Whitt1982} proposed two methods for constructing such a renewal
approximation for a general point process by matching its long-run variability,
measured through the IDC.
Each original stream is replaced by a renewal process with the same rate and
long-run variability, after which the variability of the merged stream is
obtained as a rate-weighted combination of the individual variabilities.
This approach captures asymptotic variability but neglects the detailed
temporal dependence structure of the arrivals.

\cite{Albin1984} extended this framework to queueing applications by
incorporating additional information on short-range dependence and burstiness.
Instead of relying only on the long-run variability, the method adjusts the
renewal surrogate to better reflect local correlation effects, resulting in
improved accuracy when the input processes exhibit significant dependence.
Thus, Whitt’s method provides a tractable asymptotic approximation, whereas
Albin’s extension enhances practical performance by partially accounting for
the dependence structure of the superposed arrivals.

\cite{Newell1984} showed that convergence to renewal or Poisson behavior is slow and regime-dependent, and that dependence induced by superposition is intrinsic. These renewal-based approximations fundamentally ignore autocorrelation and higher-order structure, which limits their accuracy in heavily loaded systems and renders them unreliable for network settings where correlations propagate downstream.

\cite{doi:10.1287/opre.2015.1367} model arrival streams using polyhedral uncertainty sets that bound cumulative arrivals over time, capturing both average rate and variability (including burstiness and heavy tails). Instead of specifying full probability distributions, they characterize each process through deterministic constraints inspired by limit laws. Superposition is formed by aggregating multiple arrival streams and combining their corresponding uncertainty parameters into a single uncertainty set that represents the merged flow.

MAP-based approaches offer a formally exact representation of superposed arrivals, since independent MAPs can be merged through Kronecker sums of their underlying generators \cite{Neuts1979,HeNeuts1998}. This framework has been used to analyze single-server queues with non-renewal arrivals \cite{Lucantoni1990} and to model correlated traffic streams. However, MAP-based superposition suffers from severe practical limitations: the state dimension grows multiplicatively with each merge, fitting MAPs to prescribed moment and correlation information is an ill-posed inverse problem \cite{HorvathTelek2002}, and queueing analysis with high-dimensional MAP inputs quickly becomes computationally infeasible. Consequently, MAPs are not a scalable solution for open queueing networks with repeated superposition.

Several works have therefore retained exact superposition while weakening the scope of queueing analysis. \cite{Torab2001} preserves the exact superposition of renewal processes but replaces the resulting non-renewal arrivals with renewal surrogates characterized primarily by mean rate and squared coefficient of variation, enabling approximate GI/GI/1 analysis. \cite{Wagner1994} model superposition exactly without renewal assumptions, but restricts queueing results to mean-value performance measures. In both cases, the loss of distributional information or dependence structure limits predictive accuracy and applicability.

So far, we have reviewed studies that assume given arrival streams. As discussed in the introduction, the present paper leverages the ability to merge arrival streams together with learning-based approximations of departure processes and steady-state distributions in order to analyze a broad class of open queueing networks (OQNs). As noted earlier, \cite{https://doi.org/10.1002/nav.22010} also proposes a method for merging non-renewal arrival streams and embeds this capability within a framework for analyzing OQNs. Their objective is therefore closely related to ours. For this reason, we provide a direct comparison between the two approaches (see Section~\ref{sec:questa}).

A machine-learning-based framework is proposed in~\cite{doi:10.1080/00207543.2021.1887536}, where Gaussian Process (GP) regression is used to approximate performance measures in an OQN. The model estimates both the departure process and the mean waiting time at each station, relying solely on the first two moments and the first-lag autocorrelation of the arrival streams as input descriptors. The proposed framework is flexible with respect to network structure and accommodates both superposition and flow splitting.

A limitation of that work is that inter-arrival times are generated exclusively from correlated Weibull processes. As a result, the variability structure of the input streams is effectively represented through low-order characteristics, without systematically incorporating higher-order moments. In contrast, the findings of the present study, as well as those in~\cite{SHERZER2025141}, indicate that moments beyond the second order can significantly influence queueing performance, especially in regimes where variability and temporal dependence jointly impact congestion dynamics. Since it is a black box regression model without an open source, it would be impossible to compare its results to ours. 

Collectively, the literature demonstrates that no existing superposition method is practically compatible with accurate and scalable queueing analysis: renewal approximations discard dependence, MAP-based methods are computationally prohibitive, and exact non-renewal models yield only weak analytical results. Recent learning-based approaches offer a fundamentally different alternative. In particular, \cite{SHERZER2025141} shows that steady-state queue-length distributions and departure-process statistics of $G/GI/1$ and tandem queueing systems can be accurately approximated directly from moments and autocorrelation information. Building on this framework, the present work introduces a data-driven superposition operator that maps the moments and autocorrelations of multiple non-renewal arrival streams to those of their merged process. By combining this operator with learning-based queueing analysis, steady-state probabilities in open queueing networks with superposition can be approximated without renewal assumptions or MAP representations, overcoming a longstanding limitation in non-Markovian queueing theory.

\section{Problem formulation} \label{sec:prob_form}

In this section, we formalize the proposed superposition operator. We define the superposition of two non-renewal arrival streams; extension to any $n \geq 2$ streams is obtained by sequential application of the same operator. We introduce the following notation:

\begin{itemize}
    \item $A^j_q$ denotes the $q^{\text{th}} \in \mathbb{N}$ stochastic inter-arrival time of stream $j \in \{1,2\}$.
    
    \item $A^j$ denotes the equilibrium inter-arrival time of stream $j \in \{1,2\}$.
    
    \item $S_q$ denotes the $q^{\text{th}} \in \mathbb{N}$ stochastic inter-arrival time of the superposed process obtained from streams 1 and 2.
    
    \item $S$ denotes the equilibrium inter-arrival time of the superposed process.
    
    \item $m_{A_j}(i)$ denotes the $i^{\text{th}}$ moment of the inter-arrival time distribution of stream $j \in \{1,2\}$, for $i \in \mathbb{N}$.
    
    \item $m_{S}(i)$, $\hat{m}_{S}(i)$ denotes the true and the estimated $i^{\text{th}}$ moment of $S$, respectively, for $i \in \mathbb{N}$.
    
    \item $\rho_{A_j}(k,a_1,a_2)$ denotes the $k^{\text{th}}$-lag autocorrelation structure, defined in Equation~\eqref{eq:rho_definition}, for $k,a_1,a_2 \in \mathbb{N}$.
    
    \item $\rho_{S}(k,a_1,a_2)$ and $\hat{\rho}_{S}(k,a_1,a_2)$ denotes true the estimated $k^{\text{th}}$-lag autocorrelation under the superposed process $S$, respectively, according to the defintion of Equation~\eqref{eq:rho_definition}.
\end{itemize}

The autocorrelation structure is defined as
\begin{align} \label{eq:rho_definition}
\rho_{A^j}(k,a_1,a_2)
=
\frac{
\operatorname{Cov}\left((A^j_q)^{a_1}, (A^j_{q-k})^{a_2}\right)
}{
\sqrt{
\operatorname{Var}\left((A^j_q)^{a_1}\right)
\operatorname{Var}\left((A^j_{q-k})^{a_2}\right)
}
}.
\end{align}

Our objective is to construct a mapping from the statistical descriptors of the two input arrival streams to those of their superposition.

Formally, the input to the superposition operator is:
\[
\Big(
m_{A_j}(i),
\rho_{A_j}(k,a_1,a_2)
\Big),
\quad
j \in \{1,2\},
\quad
i \leq n,
\quad
a_1,a_2 \leq n_1,
\quad
k \leq n_2,
\]
where $n$ is the number of moments used from each original stream, and $(n_1,n_2)$ determines the maximum polynomial order and lag considered in the autocorrelation structure. The output of the operator is
\[
\Big(
\hat{m}_{S}(i),
\hat{\rho}_{S}(k,a_1,a_2)
\Big),
\quad
i \leq b,
\quad
a_1,a_2 \leq b_1,
\quad
k \leq b_2,
\]
where $b$ is the target number of moments of the merged process, and $(b_1,b_2)$ specify the maximum polynomial order and lag for the merged autocorrelation structure. To ensure compatibility with the queueing analysis framework of \cite{SHERZER2025141}, we set
\[
b = 5,
\qquad
b_1 = b_2 = 2.
\]

The input parameters $(n,n_1,n_2)$ are not required to equal $(b,b_1,b_2)$; rather, they are selected to optimize predictive performance of the superposition operator.

\section{Training process} \label{sec:training_proc}

In this section, we begin by describing the procedure used to generate a broad collection of non-renewal processes for the training phase (see Section~\ref{sec:data_generation}). We then present the neural network architecture in Section~\ref{sec:network}, and conclude with a description of the customized loss function in Section~\ref{sec:loss_func}.

\subsection{Data generating}\label{sec:data_generation}

We aim to construct a flexible generation mechanism capable of producing a broad family of MAPs. In particular, the objective is to simultaneously obtain processes with rich marginal characteristics and diverse dependence structures. The variability of the marginal inter-arrival distribution is assessed through the SCV, skewness, and kurtosis, while the dependence structure is quantified by the first-lag autocorrelation.

A MAP is a versatile point process that generalizes the Poisson process by allowing dependence and variability beyond the exponential case. A MAP is characterized by two square matrices $D_0$ and $D_1$ of the same dimension $m$, where $D_0$ governs transitions of an underlying continuous-time Markov chain that do not produce arrivals, and $D_1$ governs transitions that are accompanied by an arrival. The infinitesimal generator of the background Markov chain is given by $D = D_0 + D_1$, where $D$ is a stable generator (its rows sum to zero, off-diagonal elements are nonnegative, and diagonal elements are negative). The arrival rate of the process is $\lambda = \boldsymbol{\pi} D_1 \mathbf{e}$, where $\boldsymbol{\pi}$ is the stationary distribution satisfying $\boldsymbol{\pi} (D_0 + D_1) = 0$ and $\mathbf{e}$ is a column vector of ones. Through the interaction between $D_0$ and $D_1$, a MAP can capture arbitrary inter-arrival time distributions (of phase-type) as well as positive or negative autocorrelation between successive inter-arrival times.

Uniformly sampling the entries of $D_0$ and $D_1$ subject only to feasibility constraints (i.e., nonnegative off-diagonal elements and row sums consistent with a valid generator) does not yield a representative or sufficiently rich collection of MAPs. Such naive sampling typically concentrates probability mass on a limited region of the admissible parameter space, leading to marginal inter-arrival distributions with moderate SCV and weak dependence, while extreme variability, heavy skewness, high kurtosis, or strong positive and negative autocorrelations are rarely attained. To systematically explore a broad range of behaviors, it is therefore necessary to generate MAPs from carefully designed structural families, where each predefined structure is tailored to span a specific domain in terms of SCV, skewness, and kurtosis, together with controlled levels of first-lag autocorrelation. By combining multiple such structured constructions, one can effectively cover diverse regions of the MAP space that would be inaccessible under uniform sampling.

We generate MAPs using three distinct algorithms to achieve broad coverage over both marginal distributions and dependence structures. The first method produces strongly negative autocorrelation, whereas the second induces strongly positive autocorrelation. Although these two approaches yield pronounced dependence, they typically offer limited variability in the marginal characteristics (e.g., SCV). The third method, in contrast, generates moderate levels of autocorrelation while allowing for substantially richer and more diverse marginal behavior.

Strong negative autocorrelation in a two-regime MAP is typically achieved through near-deterministic alternation between a fast and a slow regime. In this setting, short inter-arrival times are almost always followed by long ones, and vice versa. This alternating structure suppresses clustering of large values and prevents consecutive long (or consecutive short) inter-arrivals. As a result, variability is partially “balanced out” from one step to the next: a long interval is immediately compensated by a short one, which reduces dispersion over short time scales.

From a moment perspective, obtaining very large SCV requires occasional bursts of long inter-arrivals or strong tail behavior. However, strong negative autocorrelation eliminates such bursts, because long intervals cannot appear consecutively. Consequently, while mean separation between regimes can increase variability to some extent, the enforced alternation constrains the attainable second moment. This structural restriction makes it difficult to simultaneously achieve first-lag correlation and very large SCV, so strongly negative correlation typically comes with only moderate variability. We follow this line of thought when producing Algorithm~\ref{alg:map_strong_negative} in Appendix~\ref{append:gen_map}.  

Strong positive autocorrelation is obtained by introducing highly persistent regimes in the MAP. In a two-regime structure, this is achieved by setting the probability of remaining in the same regime close to one, so that the process generates long runs of short inter-arrivals (fast regime) or long inter-arrivals (slow regime). This clustering of similar values produces pronounced positive correlation. Unlike the negatively correlated case, persistence amplifies variability: consecutive long inter-arrivals increase the second moment and strengthen tail behavior. Consequently, with sufficient mean separation and high regime stickiness, strong positive correlation can naturally coexist with large SCV and substantial higher-order variability. We follow this line of thought when producing Algorithm~\ref{alg:map_strong_positive} in Appendix~\ref{append:gen_map}.  

To provide a wide range of MAPs with milder autocorrelation, we suggest the following. The generator constructs a two-regime MAP designed to produce a prescribed \emph{moderate} first-lag dependence (positive or negative) while maintaining highly variable marginal inter-arrival shapes. Two PH ``regimes'' are defined: a \emph{fast} regime with mean $\texttt{mean\_fast}$, typically implemented as an Erlang PH (hence relatively low variability), and a \emph{slow} regime with mean $\texttt{mean\_slow}$ whose tail behavior is randomized either a hyperexponential PH (large rate ratio) or a mixture of hyperexponential and Erlang components to broaden the attainable SCV, skewness, and kurtosis. Dependence is then introduced via a $2\times2$ regime-transition matrix $R$ of the form
$$
R=\begin{pmatrix} p_{\mathrm{stay}} & 1-p_{\mathrm{stay}}\\[2pt] 1-p_{\mathrm{stay}} & p_{\mathrm{stay}}\end{pmatrix},
$$
where $p_{\mathrm{stay}}$ is a smooth function of the desired $\rho_{\texttt{target}}$ and is clipped to $(0.05,0.95)$. When $p_{\mathrm{stay}}>1/2$ the process tends to persist in the same regime, yielding positive autocorrelation through clustering of short (fast) or long (slow) inter-arrival times; when $p_{\mathrm{stay}}<1/2$ the process tends to alternate between regimes, which induces negative autocorrelation by pairing short inter-arrivals with long ones. The magnitude of the autocorrelation is primarily controlled by the regime ``stickiness'' $p_{\mathrm{stay}}$ and the separation between the regime means, whereas the marginal SCV and higher standardized moments are controlled by the PH shapes within each regime.   We follow this line of thought when producing Algorithm~\ref{alg:map_medium_corr_varied_marginals} in Appendix~\ref{append:gen_map}.  

\begin{figure}
\centering
\includegraphics[scale = 0.65]{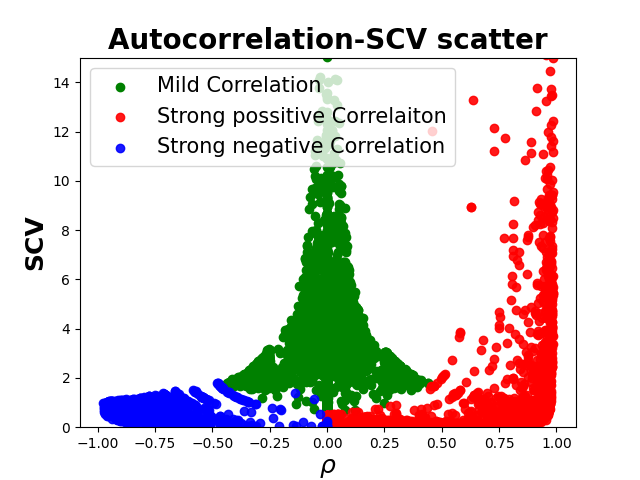}
\caption{Autocorrelation-SCV scatter.  }
\label{fig:SCV_rho_scatter1}
\end{figure}

\begin{figure}[ht]
    \centering
    \begin{subfigure}[b]{0.48\textwidth}
        \centering
        \includegraphics[width=\textwidth]{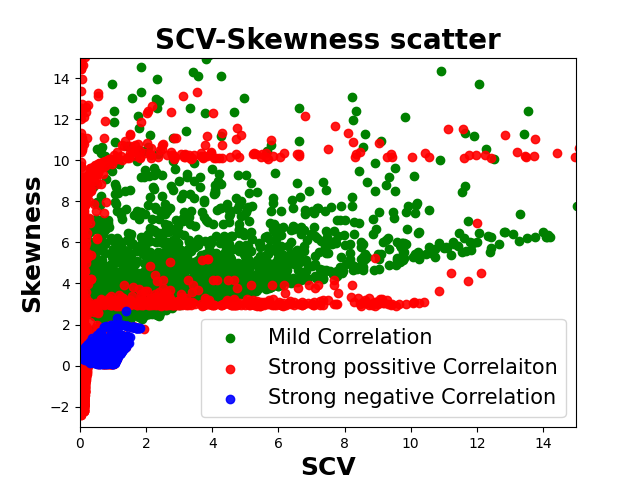}
        \caption{SCV-Skewness scatter. }
\label{fig:scv_skewness}
    \end{subfigure}
    \hfill
    \begin{subfigure}[b]{0.48\textwidth}
        \centering
        \includegraphics[width=\textwidth]{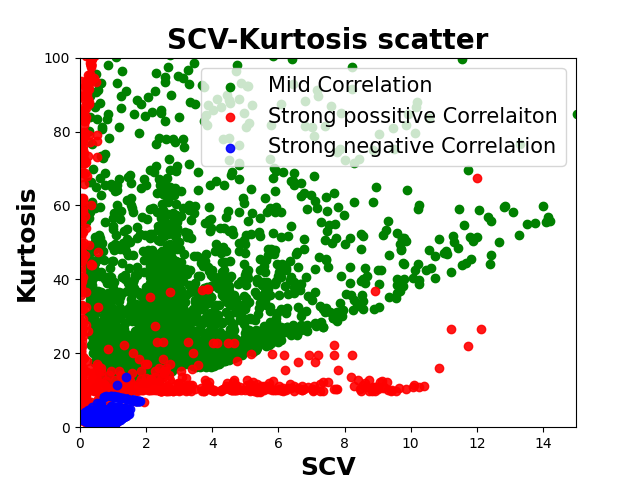}
        \caption{SCV-Kurtosis scatter.}
\label{fig:scv_kurtosis}
    \end{subfigure}
    \caption{Statistical measures of the MAP generating process.}
    \label{fig:generatingproc}
\end{figure}

We illustrate the variability generated by the three MAP construction methods in Figures~\ref{fig:SCV_rho_scatter1}, \ref{fig:scv_skewness}, and \ref{fig:scv_kurtosis}. Figure~\ref{fig:SCV_rho_scatter1} presents the joint scatter of first-lag autocorrelation and SCV across the three generators. As observed, the mild and strong-positive correlation algorithms (green and red points, respectively) produce MAPs spanning a broad range of SCV values. In contrast, the strong negative correlation generator is confined to a considerably narrower SCV domain (the blue dots). The precise ranges attained by each method are summarized in Table~\ref{tab:gen_MAP_res}. 

Figures~\ref{fig:scv_skewness} and \ref{fig:scv_kurtosis} illustrate the variability of skewness and kurtosis as functions of the SCV. We restrict the displayed MAPs to SCV values up to 15, skewness up to 50, and kurtosis up to 100, since observations beyond these ranges become increasingly sparse. The results indicate that the mild-correlation generator spans the broadest domain in terms of SCV, skewness, and kurtosis. This behavior is expected, as strong correlations typically arise from more structured MAP constructions and therefore occupy a narrower region of the feasible variability space. 

\begin{table}[!htp]\centering
\caption{Generating MAPs domain}\label{tab:gen_MAP_res}
\begin{tabular}{lrrrrrr}\toprule
Method &Discription &Autocorrelation &SCV &Skewness &Kurtosis \\ \hline
1 &Strong Negative Correlation &(-0.99,0) &(0.15, 1.81) &(0.065,2.64) &(1.09, 13.09) \\
2 &Strong Positive Correlation &(0,0.99) &(0.01,15) &(-2.41, 50) &(1.11, 100) \\
3 &Mild Correlation &(-0.43, 0.44) &(0.15, 15) &(1.74, 25.1) &(9.01,100) \\
\bottomrule
\end{tabular}
\end{table}

Once the MAP generator is defined, a single experimental instance is obtained by sampling a pair of MAPs. By construction, the generator produces processes with unit mean. Without loss of generality, we rescale the first stream such that the mean is less than 1. Consequently, during inference, time is normalized so that the MAP with the larger mean is set to one. This normalization reduces the effective training domain while preserving full generality at inference time.

\begin{remark}
We emphasize that generating the widest possible class of MAPs is not the primary objective of this work. The constructed families already span a substantial range of variability and dependence regimes, including moderate-to-high SCV values and both positive and negative correlation structures. This domain is sufficiently rich to validate the robustness and generality of the proposed neural superposition framework. Expanding the generator to cover additional extreme configurations constitutes a natural extension but is beyond the scope of the present study.
\end{remark}

\subsubsection{Labeling}\label{sec:labeling}

The labeling procedure is based on the exact superposition of MAPs. Given two independently generated MAPs, their superposition is computed using the classical Kronecker construction \cite{Neuts1979,HeNeuts1998}. 
Specifically, if the two arrival streams are represented by $(D_0^{(1)},D_1^{(1)})$ and $(D_0^{(2)},D_1^{(2)})$, the merged MAP is constructed via the standard Kronecker-sum formulation. 
This produces an exact representation of the superposed process.

In this work, MAPs of dimension up to 100 states are generated for each individual stream. 
As a result, the superposed MAP may reach dimensions up to $100 \times 100 = 10{,}000$ states. 
Such large state spaces make the computation of steady-state distributions, higher-order moments, and autocorrelation measures computationally intensive.

For each superposed MAP, we compute:
 the first $n = 10$ moments of the inter-arrival time distribution, and
     autocorrelation coefficients defined in Equation~\eqref{eq:rho_definition} 
    for $a_1,a_2 \leq 5$ and $k \leq 5$. This results in the evaluation of 10 marginal moments and 135 autocorrelation descriptors per labeled instance.

Due to the dimensionality of the superposed MAP, substantial computational resources are required. 
When the dimension of the merged MAP exceeds 200 states, computations are executed on a high-performance server equipped with an Intel Xeon Gold 6248 processor ( 2.5 GHz 1024 GB RAM). 
For smaller MAPs, computations are performed on a workstation with an Intel Core i7-14650HX processor (2.20 GHz, 32 GB RAM).

The average labeling time per instance is approximately 7.6 minutes for smaller MAPs and 8.9 minutes for larger MAPs (when used on a stronger computing power). 
These timings reflect the cost of constructing the Kronecker superposition and evaluating all required moment and autocorrelation descriptors.

\begin{remark}\label{rem:log}
 We apply a logarithmic transformation to the moments in order to reduce the dynamic range of their values. 
This follows standard practice in neural network modeling, where inputs are typically scaled to lie within comparable numerical ranges. 
As the order of the moment increases, its magnitude grows rapidly, often exponentially, so even standardized moments may exhibit substantial scale disparities. 
Moreover, relying solely on standardization increases computational overhead without fully addressing the issue of range imbalance. Such transformation was introduced in~\cite{sherzer23}, and was utilized in~\cite{SHERZER2025141, SHERZER2025889} as well.
\end{remark}

\subsubsection{Datasets: training, validation and test}\label{sec:datasets} 

For each neural network, we construct three datasets: training, validation, and testing. 
The validation set is used for hyperparameter tuning, including the selection of learning rate, network depth, and other architectural parameters. 
A complete specification of the chosen hyperparameters is provided in Appendix~\ref{append:finetune}.

The sizes of the training, validation, and test sets are 500{,}000, 10{,}000, and 10{,}000 samples, respectively.

\subsection{Network Architecture}\label{sec:network}

We implement a fully connected feed-forward neural network. 
A complete specification of the network architecture is provided in Appendix~\ref{append:architecture}. 
All models considered in this work are fully connected networks, commonly referred to as artificial neural networks (ANNs).

The choice of ANNs over Convolutional Neural Networks (CNNs) and Recurrent Neural Networks (RNNs) is motivated by the structure of the problem. 
CNNs are not well-suited to our setting due to the relatively low-dimensional input representation, which lacks spatial structure. 
RNNs, on the other hand, are designed for sequential and time-dependent data, whereas our inputs consist of stationary statistical descriptors. 
Further discussion on architectural considerations can be found in \cite{NIPS2015_b618c321,SHERZER2025889}.

\subsection{Loss function}\label{sec:loss_func}
In deep learning, model training is commonly carried out using mini-batch optimization. 
At each iteration, a subset of the full training dataset, referred to as a batch, is used to compute the gradient and update the model parameters. 
We denote the batch size by $B$, which represents the number of training samples processed before each update of the network weights. We denote the $b^{th}$ instance stochastic superposed inter-arrival time of a batch by $S^b$. 

\begin{align*}
    &Loss(\cdot) = \\ & \sum_{b=1}^B\! \left[ \! \frac{\alpha}{B} \! \left(\sum_{i=1}^n \left[log(m_{S^{b}}(i)) \!-\! log(\hat{m}_{S^{b}}(i))\right]^2 \right) \! + \!  \frac{(1\!-\!\alpha)}{B} \left(\sum_{k=1}^{n_1} \!\sum_{a_1=1}^{n_2}\!\sum_{a_2=1}^{n_2} \!\left[ \rho_{S^b}(a_1, a_2, k)\! - \!\hat{\rho}_{S^b}(a_1, a_2, k)\right]^2 \right) \right]
\end{align*}

\noindent where $0 \leq \alpha \leq 1$ is a hyperparameter controlling the relative weight of the two components. 
As noted in Remark~\ref{rem:log}, the network is trained on the logarithms of the moments, which explains the presence of the log operator in the loss function. 

The loss function consists of two components corresponding to the moments and the autocorrelation descriptors. 
For each component, we compute the mean squared error (MSE), and the overall loss is obtained as their weighted combination via $\alpha$.

\section{Experiments} \label{sec:experiments}

We conduct two sets of experiments to evaluate the proposed superposition operator. 
The first experiment examines the impact of the number of input descriptors on prediction accuracy. 
The second experiment evaluates the robustness of the neural network across structurally different arrival-process regimes.

Throughout the experiments, the target quantities are the first $b=5$ moments and eight autocorrelation values corresponding to lags up to $b_2=2$ and polynomial order up to $b_1=2$, as stated in Section~\ref{sec:prob_form}.

There are a total of 8 autocorrelation values. From here forward, we present the accuracy results in the following order:
$\hat{\rho}_{S}(1,1,1)$, $\hat{\rho}_{S}(1,1,2)$, 
$\hat{\rho}_{S}(1,2,1)$,
$\hat{\rho}_{S}(1,2,2)$,
$\hat{\rho}_{S}(2,1,1)$, $\hat{\rho}_{S}(2,1,2)$, 
$\hat{\rho}_{S}(2,2,1)$
$\hat{\rho}_{S}(2,2,2)$. 

\subsection{Accuracy Metrics}

To evaluate the accuracy of the predicted moments, we use the Mean Absolute Percentage Error (MAPE), a standard metric in the queueing-learning literature and also used in \cite{SHERZER2025141} to evaluate moment accuracy. 
MAPE is defined as
\begin{align*}
\text{MAPE}
=
\frac{100}{v}
\sum_{l=1}^{v}
\left|
\frac{y_l - \hat{y}_l}{y_l}
\right|,
\end{align*}
where $y_l$ and $\hat{y}_l$ denote the true and predicted values of the $l^{\text{th}}$ sample, respectively, and $v$ is the sample size. 
In our setting, $y_l$ and $\hat{y}_l$ correspond to the true and predicted values of a specific moment of the superposed inter-arrival time distribution. 
The MAPE is computed separately for each moment. 
Since the first moment (the mean) of the superposed process is known exactly from the input streams, the accuracy evaluation is performed only for the second through fifth moments.

While MAPE is suitable for moment evaluation, it is less appropriate for autocorrelation descriptors. 
Autocorrelation values are typically small in magnitude, and when the true value is close to zero, even small absolute deviations may produce disproportionately large percentage errors. 
This can lead to misleading interpretations of model performance by overemphasizing minor deviations.

To avoid this issue, we evaluate autocorrelation predictions using the Mean Absolute Error (MAE), defined as
MAE is defined as
\begin{align*}
\text{MAE}
=
\frac{1}{v}
\sum_{i=1}^{v}
\left| y_i - \hat{y}_i \right|,
\end{align*}
where $y_i$ and $\hat{y}_i$ denote the true and predicted values of the $i^{\text{th}}$ sample, respectively, and $v$ is the sample size. 
In this study, the metric is evaluated separately for each of the eight autocorrelation descriptors defined above, with $y_i$ and $\hat{y}_i$ representing the true and predicted values of the corresponding descriptor.
 
The MAE directly measures absolute deviation without normalizing by the magnitude of the true value, making it more appropriate for small-scale descriptors such as autocorrelations.

\subsection{Experiment 1: Sensitivity to Input Descriptor Dimension}

In the first experiment, we investigate how prediction accuracy depends on the number of statistical descriptors used from the original arrival streams.

Let $n$ denote the number of moments extracted from each input stream, and let $n_1$ and $n_2$ denote the maximum lag and maximum polynomial order used in the autocorrelation descriptors, respectively. We evaluate all combinations satisfying
\[
2 \leq n \leq 10, 
\qquad
1 \leq n_1 \leq 5,
\qquad
1 \leq n_2 \leq 5.
\]

For each configuration $(n,n_1,n_2)$, the neural network is trained and evaluated, and we report the overall average PARE for the five target moments and the overall average of the eight target autocorrelation values. This experiment quantifies the marginal contribution of higher-order moments and extended correlation structure to the accuracy of the superposition operator.

\subsection{Experiment 2: Structural Robustness Across Regimes}

The second experiment evaluates the NN’s robustness across structurally distinct arrival-process regimes.

As described earlier, when generating paired streams, we scale them so that the stream with the larger mean has a first moment equal to 1. 
Hence, the second stream has a mean strictly less than 1. We partition the test data according to:

\begin{enumerate}
\item[(a)] \textbf{SCV-Based Partition.}  
For each stream, we classify the squared coefficient of variation (SCV) into two categories: SCV $< 3$ and SCV $\geq 3$. In addition, we partition the instances according to whether the ratio of the mean arrival rates of the two streams is smaller than $0.5$ or greater than $0.5$. These criteria jointly define $2 \times 2 \times 2 = 8$ distinct regimes.

The secondary partition based on the mean ratio is introduced solely to control for potential confounding effects arising from differences in stream arrival rates.

\item[(b)] \textbf{Autocorrelation-Based Partition.}  
For each stream, the first-lag autocorrelation is partitioned into four ranges:
\[
(-1,-0.25), \quad
(-0.25,0), \quad
(0,0.25), \quad
(0.25,1),
\]
representing strong negative, weak negative, weak positive, and strong positive dependence, respectively. 
We again partition according to whether the mean ratio is smaller than 0.5 or greater than 0.5. 
This results in $4 \times 4 \times 2 = 32$ regimes.
\end{enumerate}

For each of the 8 SCV-based regimes and each of the 32 autocorrelation-based regimes, we evaluate the model on 1{,}000 samples. 
The total test size for this experiment is therefore 40{,}000 samples.

For every regime, we report:
\begin{itemize}
    \item the average PARE for the five target moments,
    \item the average MAE for the eight target autocorrelation descriptors.
\end{itemize}

\begin{remark}
The SCV-based and autocorrelation-based partitions are not combined into a single joint classification. 
Although a fully joint partition would provide a finer-grained breakdown, it would generate 
$8 \times 32 = 256$ distinct subgroups. 
Such a high degree of fragmentation would substantially reduce interpretability and complicate the presentation of results. 
By analyzing variability-driven and dependence-driven regimes separately, we maintain clarity,  and still provide a systematic robustness assessment across both dimensions.
\end{remark}

This experiment provides a detailed characterization of model performance across heterogeneous variability and dependence structures.

For the sake of comparison, we also compute the superposed SCV of 3 different approximation methods from the literature and then extract the 2$^{nd}$ moment PARE, which is compared to our NN approximations. The first two methods originate from~\cite{Whitt1982}. As mentioned above, two renewal approximations for general point processes, commonly referred to as the renewal (interval) method and the asymptotic (variance–time) method, are proposed. The first method will be referred to as 'Whitt R', and the second as 'Whitt A', where 'R' and 'A' denote renewal and asymptotic, respectively. The third method is given in~\cite {Albin1984} and is referred to in the results as Albin.   While~\cite{WHITT201999} also provide superposition abilites they do not compute the merged moments nor autocorrelations hence they cannot be compared at this stage. However, thier results will be used for compestin in Section~\ref{sec:questa} when analyzing queueing systems. 

\section{Results} \label{sec:results}

We commence with the moment and autocorrelation analysis in Section~\ref{sec:mom_anal} followed by the accuracy performance in Section~\ref{sec:acc_res}. We end this section with presenting the inference runtimes in Section~\ref{sec:runtimes}.

\subsection{Moment and autocorrelation analysis}   \label{sec:mom_anal}

\begin{figure}[ht!]
    \centering

    \begin{subfigure}{0.27\textwidth}
        \centering
        \includegraphics[width=\textwidth]{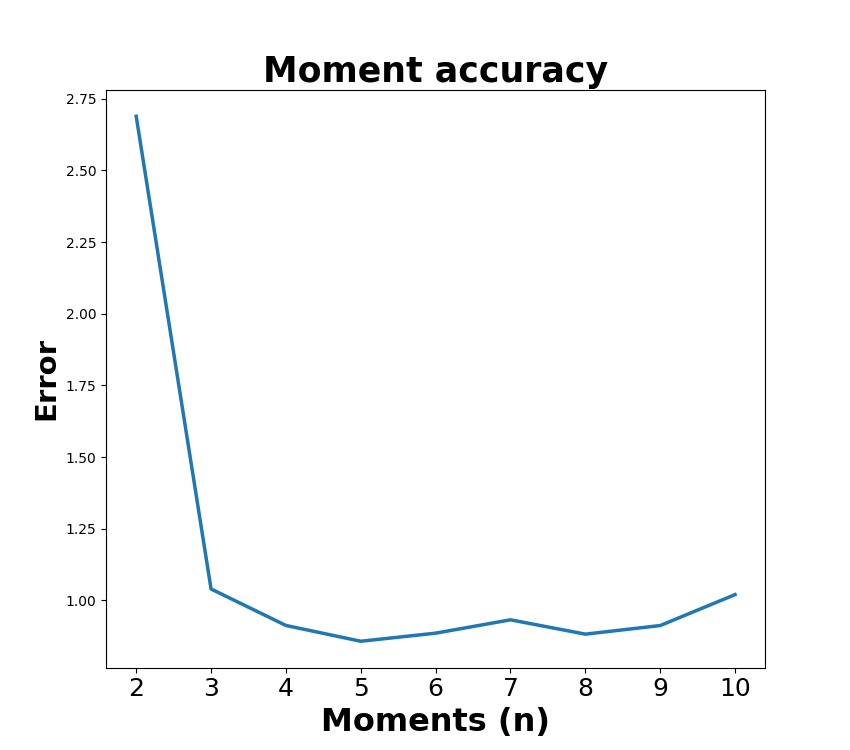}
        \caption{Impact of $n$.}
        \label{fig:mom_anal_moms}
    \end{subfigure}
    \hfill
    \begin{subfigure}{0.27\textwidth}
        \centering
        \includegraphics[width=\textwidth]{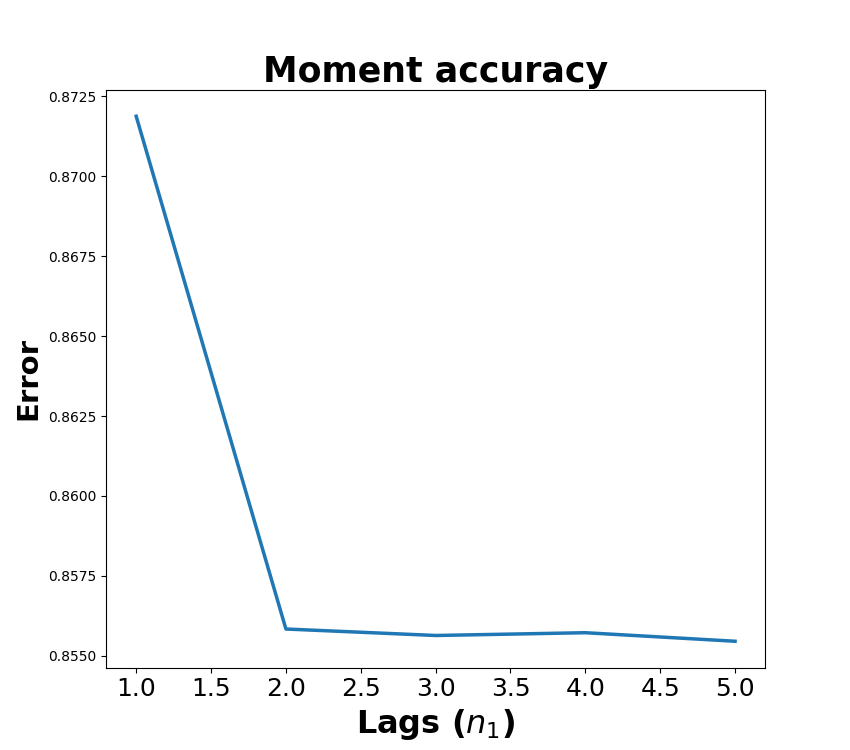}
        \caption{Impact of $n_1$.}
        \label{fig:mom_anal_lags}
    \end{subfigure}
    \hfill
    \begin{subfigure}{0.27\textwidth}
        \centering  \includegraphics[width=\textwidth]{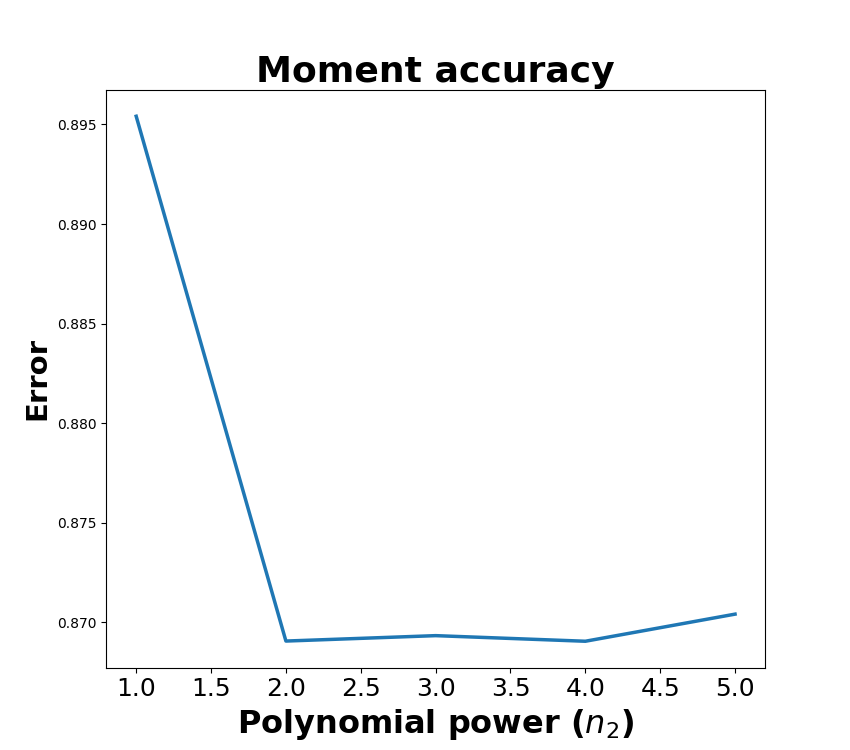}
        \caption{Impact of $n_2$.}
        \label{fig:mom_anal_power}
    \end{subfigure}

    \caption{Moment analysis}
    \label{fig:mom_anal}
\end{figure}

The results indicate that the superposition operator is identifiable from a compact statistical descriptor. 
As shown in Fig.~\ref{fig:mom_anal_moms}, the best performance is obtained at $n=5$, after which additional moments do not improve accuracy. 
Regarding temporal dependence, Fig.~\ref{fig:mom_anal_lags} shows that $n_1=3$ provides the lowest error, although only marginally better than $n_1=2$. 
Finally, the nonlinear correlation transformation saturates already at quadratic order, with the best accuracy achieved at $n_2=2$ (Fig.~\ref{fig:mom_anal_power}). 
Overall, the mapping from the two input processes to the superposed moments depends mainly on moderate-order variability and short-range dependence, and does not require higher-order statistical detail.

\begin{figure}[ht!]
    \centering

    \begin{subfigure}{0.27\textwidth}
        \centering
        \includegraphics[width=\textwidth]{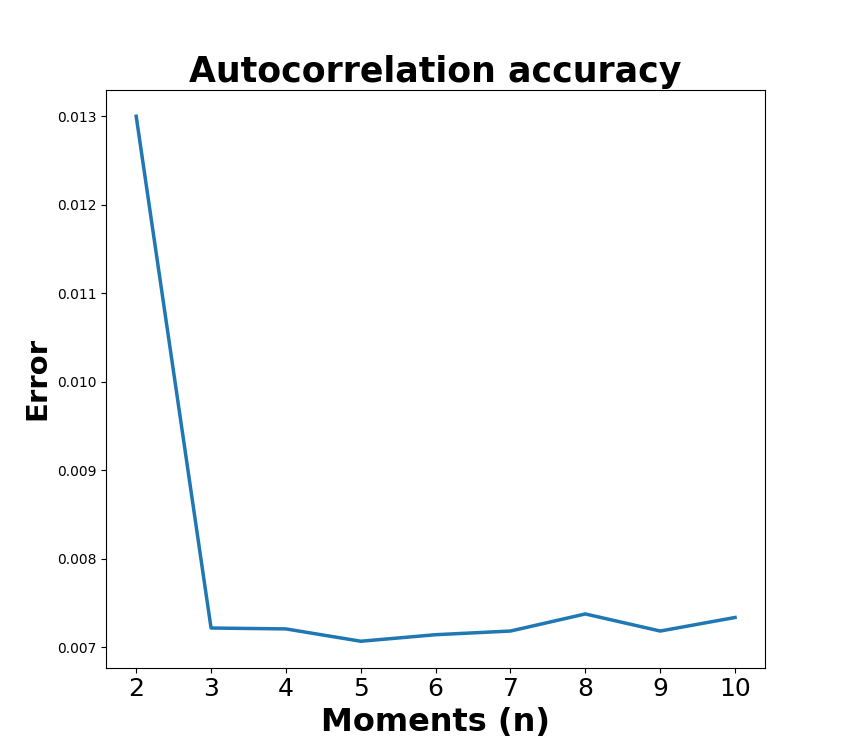}
        \caption{Impact of $n$.}
        \label{fig:cor_anal_moms}
    \end{subfigure}
    \hfill
    \begin{subfigure}{0.27\textwidth}
        \centering
        \includegraphics[width=\textwidth]{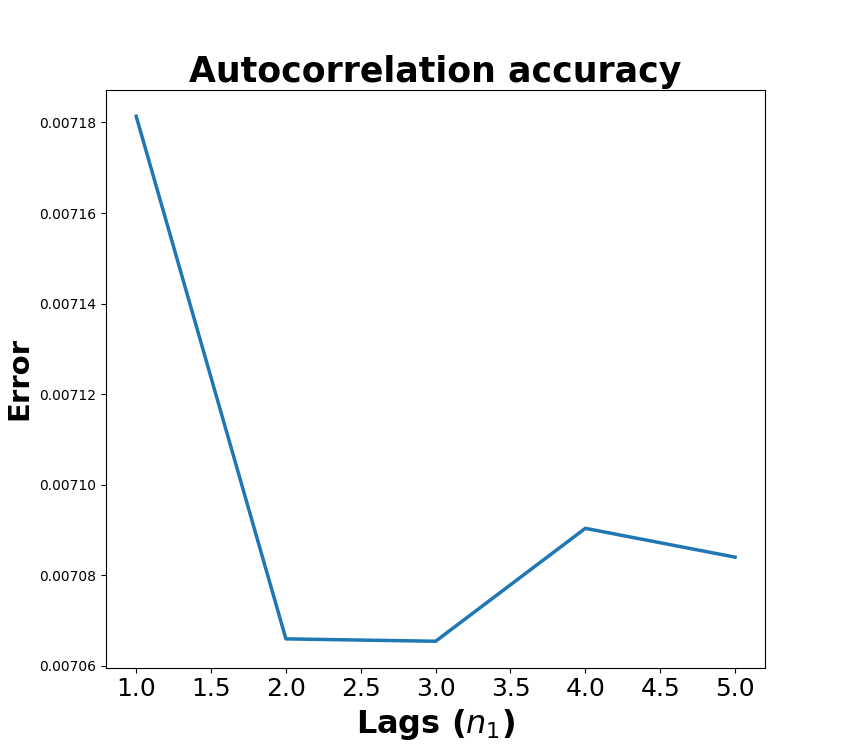}
        \caption{Impact of $n_1$.}
        \label{fig:cor_anal_lags}
    \end{subfigure}
    \hfill
    \begin{subfigure}{0.27\textwidth}
        \centering  \includegraphics[width=\textwidth]{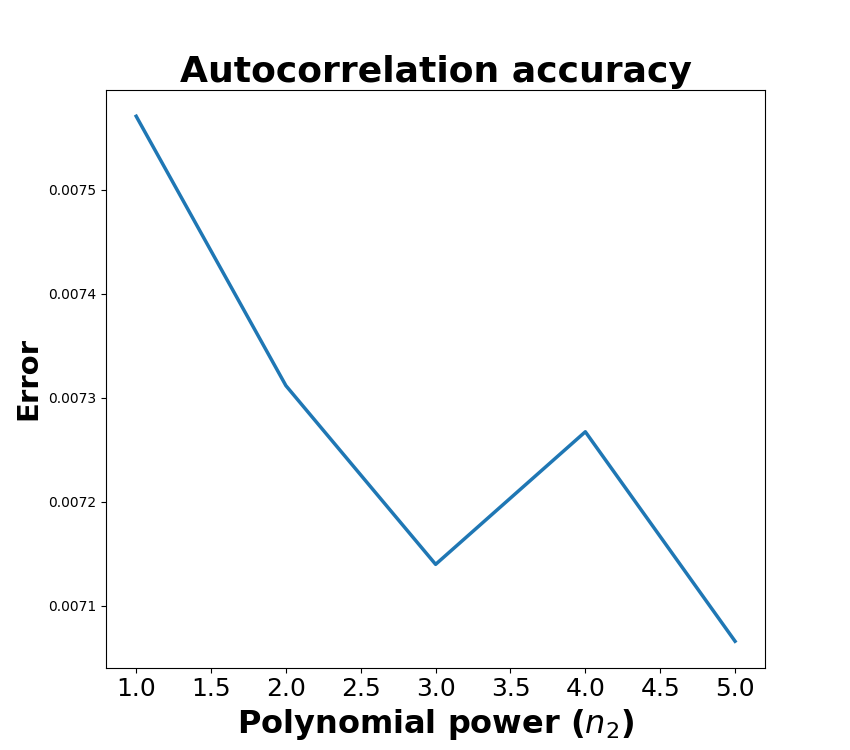}
        \caption{Impact of $n_2$.}
        \label{fig:cor_anal_power}
    \end{subfigure}

    \caption{Autocorrelation analysis}
    \label{fig:cor_anal}
\end{figure}

Figure~\ref{fig:cor_anal_moms} examines the impact of the number of input moments $n$ on the accuracy of the predicted autocorrelation descriptors. The error decreases sharply when increasing $n$ from $2$ to $4$–$5$, after which additional moments provide negligible improvement. The best overall performance is attained at $n=5$, indicating that moderate-order moment information is sufficient to reconstruct the dependence structure of the merged process.

Figure~\ref{fig:cor_anal_lags} evaluates the effect of the maximum lag parameter $n_1$. The prediction error drops substantially when increasing $n_1$ from $1$ to $2$, and reaches its minimum at $n_1=3$. However, the improvement from $n_1=2$ to $n_1=3$ is only marginal, suggesting that short-range dependence dominates the superposition mapping and that higher lags contribute limited additional information.

Figure~\ref{fig:cor_anal_power} studies the influence of the polynomial order $n_2$. The results show that accuracy improves significantly when moving from linear to quadratic terms, while higher-order polynomial powers do not yield further gains. The optimal configuration is obtained at $n_2=2$, indicating that quadratic transformations are sufficient to capture the nonlinear interaction between the two input streams.

Overall, the autocorrelation analysis confirms that the superposition operator primarily depends on moderate-order moments and short-range dependence. Increasing descriptor complexity beyond $(n,n_1,n_2)=(5,3,2)$ does not materially enhance predictive accuracy, demonstrating that the mapping is identifiable from a compact statistical representation.

\subsection{Accuracy} \label{sec:acc_res}

Tables~\ref{tab:mom_acc_scv}-\ref{tab:mom_acc_rho} report the prediction accuracy of the proposed neural superposition operator across heterogeneous arrival-process regimes.
For each regime, we present the PARE of the first 2-5 moments produced by the NN and, for comparison, the PARE of the second moment obtained using the classical approximations of
Whitt (renewal and asymptotic) and Albin.

Across all regimes, the NN maintains consistently low errors for all five moments.
In the SCV-based partition (Table~\ref{tab:mom_acc_scv}), the second-moment PARE of the proposed method ranges roughly
between $0.65\%$ and $2.73\%$, while higher moments remain below approximately $3.5\%$.
In the autocorrelation-based partition (Table~\ref{tab:mom_acc_rho}), the second-moment error varies between
$0.47\%$ and about $2.1\%$ across all 32 dependence regimes.
These results indicate stable predictive performance even when both variability and correlation
structures vary substantially.

A clear pattern emerges: the prediction accuracy slightly deteriorates when both streams exhibit
high variability or strong dependence, yet the error remains within a narrow and practically small range.
This demonstrates the robustness of the learned mapping across structurally different input processes.

The last three columns of Tables~\ref{tab:mom_acc_scv}-\ref{tab:mom_acc_rho} compare the neural approximation of the second moment with
Whitt's renewal approximation (Whitt~R), Whitt's asymptotic approximation (Whitt~A), and Albin's method.

Across all regimes, the neural approach improves accuracy by orders of magnitude.
In Table~\ref{tab:mom_acc_scv}, Whitt's methods typically produce errors ranging from about $30\%$ up to several hundred percent,
and exceeding $3000\%$ when both streams are highly variable, whereas the neural approximation remains below $3\%$.
Albin's method significantly improves over Whitt's approximations but still produces errors between roughly
$10\%$ and $130\%$, remaining an order of magnitude larger than the neural approach.

The same phenomenon is observed in the dependence-based partition (Table~\ref{tab:mom_acc_rho}).
Whitt's methods again exhibit large errors (often tens to hundreds of percent and occasionally above $900\%$),
while Albin's approximation ranges approximately between $7\%$ and $76\%$.
In contrast, the neural model consistently maintains errors below about $2\%$ across all 32 regimes.

The results show that renewal-based approximations capture only coarse long-run variability and therefore
fail in regimes with significant dependence or high variability.
Albin's method partially corrects this limitation but still lacks sufficient flexibility to describe the full
interaction between streams.
The neural superposition operator, by learning the mapping directly from statistical descriptors,
accurately captures these interactions, leading to uniformly low second-moment errors across all regimes.

Overall, the proposed method provides stable and accurate moment predictions across both variability-driven
and correlation-driven regimes, substantially outperforming classical superposition approximations.

\begin{table}[!htp]\centering
\caption{Moment accuracy by SCV}\label{tab:mom_acc_scv}
\begin{tabular}{|l|r|r|r|r|r|r|r|r|r|r|r|}\toprule
&\multicolumn{2}{c}{SCV } & &\multicolumn{4}{|c|}{PARE} &\multicolumn{3}{|c|}{Second moment PARE} \\ \hline
&Stream 1 &Stream 2 &Moment ratio &2 &3 &4 &5 &Whitt R &Whitt A &Albin \\ \hline
1 &$<3$ &$<3$ &$<0.5$ &0.66 &0.76 &0.94 &1.27 &50.37 &50.37 &13.54 \\ \hline
2 &$<3$ &$<3$ &$>0.5$ &0.65 &0.79 &1.01 &1.32 &62.94 &62.94 &11.9 \\ \hline
3 &$<3$ &$>3$ &$<0.5$ &1.44 &2 &2.38 &2.85 &28.96 &28.96 &10.43 \\ \hline
4 &$<3$ &$>3$ &$>0.5$ &1.99 &2.45 &2.8 &3.32 &53.85 &53.85 &17.6 \\ \hline
5 &$>3$ &$<3$ &$<0.5$ &2.38 &2.41 &2.35 &2.92 &359.73 &359.73 &108.97 \\ \hline
6 &$>3$ &$<3$ &$>0.5$ &2.28 &2.57 &2.73 &3.16 &311.03 &311.03 &118.78 \\ \hline
7 &$>3$ &$>3$ &$<0.5$ &2.73 &3.16 &3.27 &3.5 &3908.75 &390.75 &130.69 \\ \hline
8 &$>3$ &$>3$ &$>0.5$ &2.64 &2.52 &2.62 &3.09 &123.58 &123.58 &38.77 \\ \hline
\bottomrule
\end{tabular}
\end{table}

\begin{table}[!htp]\centering
\caption{Moment accuracy by first lag autocorrelation}\label{tab:mom_acc_rho}
\begin{tabular}{|l|r|r|r|r|r|r|r|r|r|r|r|}\toprule
&\multicolumn{2}{|c|}{$\rho$} & &\multicolumn{4}{|c|}{PARE} &\multicolumn{3}{|c|}{Second moment PARE} \\ \toprule
&Stream 1 &Stream 2 &Moment ratio &2 &3 &4 &5 &Whitt R &Whitt A &Albin \\ \hline
1 &$<-0.25$ &$<-0.25$ &$<0.5$ &0.47 &0.54 &0.61 &0.71 &44.73 &44.73 &8.61 \\ \hline
2 &$<-0.25$ &$<-0.25$ &$>0.5$ &0.47 &0.55 &0.72 &0.81 &63.96 &63.96 &12.22 \\ \hline
3 &$<-0.25$ &$(-0.25,0)$ &$<0.5$ &0.89 &1.22 &1.47 &1.75 &30.42 &30.42 &7.09 \\ \hline
4 &$<-0.25$ &$(-0.25,0)$ &$>0.5$ &1.41 &1.70 &1.98 &2.26 &53.85 &53.85 &11.93 \\ \hline
5 &$<-0.25$ &$(0,0.25)$ &$<0.5$ &0.85 &1.12 &1.40 &1.70 &47.26 &47.26 &11.29 \\ \hline
6 &$<-0.25$ &$(0,0.25)$ &$>0.5$ &1.26 &1.53 &1.80 &2.14 &64.54 &64.54 &13.72 \\ \hline
7 &$<-0.25$ &$(0.25,1)$ &$<0.5$ &0.72 &0.79 &0.97 &1.31 &56.43 &56.43 &12.46 \\ \hline
8 &$<-0.25$ &$(0.25,1)$ &$>0.5$ &0.75 &0.90 &1.11 &1.44 &68.29 &68.29 &13.82 \\ \hline
9 &$(-0.25,0)$ &$<-0.25$ &$<0.5$ &1.20 &1.17 &1.12 &1.48 &163.21 &163.21 &44.12 \\ \hline
10 &$(-0.25,0)$ &$<-0.25$ &$>0.5$ &1.26 &1.34 &1.44 &1.71 &142.98 &142.98 &44.18 \\ \hline
11 &$(-0.25,0)$ &$(-0.25,0)$ &$<0.5$ &2.06 &2.33 &2.44 &2.69 &81.34 &81.34 &22.71 \\ \hline
12 &$(-0.25,0)$ &$(-0.25,0)$ &$>0.5$ &1.93 &1.92 &2.26 &2.78 &147.48 &147.48 &43.07 \\ \hline
13 &$(-0.25,0)$ &$(0,0.25)$ &$<0.5$ &1.91 &2.30 &2.54 &2.94 &86.91 &86.91 &26.40 \\ \hline
14 &$(-0.25,0)$ &$(0,0.25)$ &$>0.5$ &2.03 &2.04 &2.18 &2.74 &115.00 &115.00 &31.94 \\ \hline
15 &$(-0.25,0)$ &$(0.25,1)$ &$<0.5$ &1.47 &1.56 &1.70 &2.18 &188.71 &188.71 &58.97 \\ \hline
16 &$(-0.25,0)$ &$(0.25,1)$ &$>0.5$ &1.47 &1.72 &1.90 &2.22 &132.55 &132.55 &47.21 \\ \hline
17 &$(0,0.25)$ &$<-0.25$ &$<0.5$ &1.48 &1.53 &1.60 &2.09 &180.60 &180.60 &50.34 \\ \hline
18 &$(0,0.25)$ &$<-0.25$ &$>0.5$ &1.34 &1.44 &1.60 &1.90 &216.82 &216.82 &76.18 \\ \hline
19 &$(0,0.25)$ &$(-0.25,0)$ &$<0.5$ &2.10 &2.47 &2.53 &2.72 &962.00 &951.00 &323.80 \\ \hline
20 &$(0,0.25)$ &$(-0.25,0)$ &$>0.5$ &1.94 &2.08 &2.27 &2.74 &118.84 &118.84 &34.99 \\ \hline
21 &$(0,0.25)$ &$(0,0.25)$ &$<0.5$ &2.03 &2.40 &2.61 &2.94 &73.26 &73.26 &22.25 \\ \hline
22 &$(0,0.25)$ &$(0,0.25)$ &$>0.5$ &2.03 &2.13 &2.28 &2.76 &118.60 &118.60 &34.21 \\ \hline
23 &$(0,0.25)$ &$(0.25,1)$ &$<0.5$ &1.45 &1.64 &1.79 &2.42 &147.62 &147.62 &51.97 \\ \hline
24 &$(0,0.25)$ &$(0.25,1)$ &$>0.5$ &1.42 &1.83 &2.17 &2.68 &125.22 &125.22 &56.49 \\ \hline
25 &$(0.25,1)$ &$<-0.25$ &$<0.5$ &0.95 &1.09 &1.20 &1.55 &56.46 &56.46 &18.08 \\ \hline
26 &$(0.25,1)$ &$<-0.25$ &$>0.5$ &0.94 &1.03 &1.17 &1.56 &68.16 &68.16 &22.58 \\ \hline
27 &$(0.25,1)$ &$(-0.25,0)$ &$<0.5$ &1.52 &1.79 &2.08 &2.47 &49.40 &49.40 &16.60 \\ \hline
28 &$(0.25,1)$ &$(-0.25,0)$ &$>0.5$ &1.87 &2.19 &2.52 &2.83 &75.83 &75.83 &23.03 \\ \hline
29 &$(0.25,1)$ &$(0,0.25)$ &$<0.5$ &1.42 &1.80 &2.09 &2.66 &70.63 &70.63 &21.65 \\ \hline
30 &$(0.25,1)$ &$(0,0.25)$ &$>0.5$ &1.49 &1.73 &2.02 &2.36 &69.99 &69.99 &18.59 \\ \hline
31 &$(0.25,1)$ &$(0.25,1)$ &$<0.5$ &1.51 &1.93 &2.21 &2.95 &74.90 &74.90 &26.24 \\\hline
32 &$(0.25,1)$ &$(0.25,1)$ &$>0.5$ &1.61 &1.77 &2.13 &2.86 &66.38 &66.38 &27.89 \\ \hline
\bottomrule
\end{tabular}
\end{table}

Tables~\ref{tab:corr_acc_scv} and~\ref{tab:corr_acc_rho} report the mean absolute error (MAE)
of the eight predicted autocorrelation descriptors across variability-based and dependence-based regimes,
respectively.

In Table~\ref{tab:corr_acc_scv}, the prediction errors remain uniformly small across all eight regimes.
For all autocorrelation entries, the MAE typically lies between $0.004$ and $0.013$, with the
largest observed values below $0.016$.
Even when both streams exhibit high variability, the error increases only marginally.
No regime shows a qualitative degradation of performance, indicating that the NN captures the
correlation structure independently of the variability level of the individual streams.

Table~\ref{tab:corr_acc_rho} evaluates performance across 32 dependence configurations.
The MAE remains extremely small across all combinations of negative and positive correlations.
For weak and moderate dependence regimes, the errors typically fall in the range $0.002$–$0.008$.
Even under strong positive dependence in both streams, the error remains bounded below $0.032$.
Hence, the predictive accuracy degrades smoothly as dependence strengthens but never exhibits instability.

Across both partitions, the autocorrelation errors are two orders of magnitude smaller than the scale
of the correlations themselves, indicating that the learned operator accurately reconstructs the
temporal dependence structure of the superposed process.
Importantly, accuracy depends more on the strength of dependence than on variability:
large SCV values alone do not significantly affect prediction quality,
whereas strong simultaneous correlations produce only moderate and controlled error increases.

These results demonstrate that the neural superposition operator reliably preserves
short-range dependence information, a feature absent from classical renewal-based superposition
approximations.

\begin{table}[!htp]\centering
\caption{Autocorrelation accuracy by SCV}\label{tab:corr_acc_scv}
\begin{tabular}{|l|r|r|r|r|r|r|r|r|r|r|r|r|}\toprule
&\multicolumn{2}{|c|}{SCV } & &\multicolumn{8}{|c|}{MAE} \\ \hline
&Stream 1 &Stream 2 &Moment ratio &1 &2 &3 &4 &5 &6 &7 &8 \\ \hline
1 &$<3$ &$<3$ &$<0.5$ &0.005 &0.004 &0.004 &0.005 &0.008 &0.006 &0.006 &0.009 \\ \hline
2 &$<3$ &$<3$ &$>0.5$ &0.006 &0.004 &0.004 &0.005 &0.008 &0.007 &0.007 &0.009 \\ \hline
3 &$<3$ &$>3$ &$<0.5$ &0.009 &0.005 &0.005 &0.006 &0.011 &0.008 &0.008 &0.009 \\ \hline
4 &$<3$ &$>3$ &$>0.5$ &0.007 &0.005 &0.005 &0.006 &0.010 &0.008 &0.008 &0.010 \\ \hline
5 &$>3$ &$<3$ &$<0.5$ &0.009 &0.007 &0.007 &0.008 &0.013 &0.011 &0.011 &0.014 \\ \hline
6 &$>3$ &$<3$ &$>0.5$ &0.010 &0.009 &0.009 &0.011 &0.013 &0.013 &0.013 &0.016 \\ \hline
7 &$>3$ &$>3$ &$<0.5$ &0.007 &0.005 &0.005 &0.005 &0.008 &0.006 &0.006 &0.007 \\ \hline
8 &$>3$ &$>3$ &$>0.5$ &0.007 &0.005 &0.005 &0.005 &0.008 &0.006 &0.006 &0.007 \\ \hline
\bottomrule
\end{tabular}
\end{table}

\begin{table}[!htp]\centering
\caption{Autocorrelation accuracy by first lag autocorrelation}\label{tab:corr_acc_rho}
\begin{tabular}{|l|r|r|r|r|r|r|r|r|r|r|r|r|}\toprule
&\multicolumn{2}{|c|}{$\rho$} & &\multicolumn{8}{|c|}{MAE} \\  \hline
&Stream 1 &Stream 2 &Moment ratio &1 &2 &3 &4 &5 &6 &7 &8 \\ \hline
1 &$<-0.25$ &$<-0.25$ &$<0.5$ &0.003 &0.002 &0.002 &0.002 &0.006 &0.005 &0.005 &0.006 \\ \hline
2 &$<-0.25$ &$<-0.25$ &$>0.5$ &0.003 &0.002 &0.002 &0.003 &0.006 &0.005 &0.005 &0.006 \\ \hline
3 &$<-0.25$ &$(-0.25,0)$ &$<0.5$ &0.003 &0.002 &0.002 &0.002 &0.006 &0.004 &0.004 &0.004 \\ \hline
4 &$<-0.25$ &$(-0.25,0)$ &$>0.5$ &0.003 &0.002 &0.002 &0.002 &0.006 &0.005 &0.005 &0.005 \\ \hline
5 &$<-0.25$ &$(0,0.25)$ &$<0.5$ &0.003 &0.002 &0.002 &0.002 &0.006 &0.005 &0.005 &0.005 \\ \hline
6 &$<-0.25$ &$(0,0.25)$ &$>0.5$ &0.004 &0.002 &0.002 &0.003 &0.006 &0.005 &0.005 &0.007 \\ \hline
7 &$<-0.25$ &$(0.25,1)$ &$<0.5$ &0.005 &0.003 &0.003 &0.004 &0.010 &0.008 &0.008 &0.009 \\ \hline
8 &$<-0.25$ &$(0.25,1)$ &$>0.5$ &0.007 &0.004 &0.004 &0.005 &0.009 &0.009 &0.009 &0.012 \\ \hline
9 &$(-0.25,0)$ &$<-0.25$ &$<0.5$ &0.004 &0.003 &0.003 &0.003 &0.007 &0.005 &0.005 &0.006 \\ \hline
10 &$(-0.25,0)$ &$<-0.25$ &$>0.5$ &0.004 &0.003 &0.003 &0.003 &0.006 &0.006 &0.006 &0.007 \\ \hline
11 &$(-0.25,0)$ &$(-0.25,0)$ &$<0.5$ &0.004 &0.003 &0.003 &0.003 &0.005 &0.004 &0.004 &0.004 \\ \hline
12 &$(-0.25,0)$ &$(-0.25,0)$ &$>0.5$ &0.005 &0.003 &0.004 &0.003 &0.005 &0.004 &0.005 &0.005 \\ \hline
13 &$(-0.25,0)$ &$(0,0.25)$ &$<0.5$ &0.005 &0.004 &0.004 &0.004 &0.006 &0.005 &0.005 &0.006 \\ \hline
14 &$(-0.25,0)$ &$(0,0.25)$ &$>0.5$ &0.006 &0.005 &0.005 &0.005 &0.007 &0.006 &0.006 &0.006 \\ \hline
15 &$(-0.25,0)$ &$(0.25,1)$ &$<0.5$ &0.007 &0.005 &0.006 &0.006 &0.009 &0.007 &0.007 &0.007 \\ \hline
16 &$(-0.25,0)$ &$(0.25,1)$ &$>0.5$ &0.008 &0.007 &0.006 &0.008 &0.010 &0.009 &0.009 &0.010 \\ \hline
17 &$(0,0.25)$ &$<-0.25$ &$<0.5$ &0.005 &0.003 &0.003 &0.004 &0.008 &0.007 &0.006 &0.008 \\ \hline
18 &$(0,0.25)$ &$<-0.25$ &$>0.5$ &0.005 &0.003 &0.003 &0.004 &0.007 &0.007 &0.007 &0.009 \\ \hline
19 &$(0,0.25)$ &$(-0.25,0)$ &$<0.5$ &0.005 &0.004 &0.004 &0.004 &0.006 &0.004 &0.004 &0.004 \\ \hline
20 &$(0,0.25)$ &$(-0.25,0)$ &$>0.5$ &0.006 &0.004 &0.004 &0.004 &0.007 &0.005 &0.005 &0.005 \\ \hline
21 &$(0,0.25)$ &$(0,0.25)$ &$<0.5$ &0.005 &0.004 &0.004 &0.004 &0.005 &0.004 &0.004 &0.006 \\ \hline
22 &$(0,0.25)$ &$(0,0.25)$ &$>0.5$ &0.006 &0.005 &0.005 &0.005 &0.007 &0.006 &0.006 &0.007 \\ \hline
23 &$(0,0.25)$ &$(0.25,1)$ &$<0.5$ &0.008 &0.006 &0.006 &0.008 &0.010 &0.008 &0.008 &0.011 \\ \hline
24 &$(0,0.25)$ &$(0.25,1)$ &$>0.5$ &0.009 &0.008 &0.008 &0.010 &0.013 &0.010 &0.010 &0.013 \\ \hline
25 &$(0.25,1)$ &$<-0.25$ &$<0.5$ &0.009 &0.006 &0.006 &0.007 &0.015 &0.011 &0.011 &0.018 \\ \hline
26 &$(0.25,1)$ &$<-0.25$ &$>0.5$ &0.007 &0.005 &0.005 &0.005 &0.012 &0.010 &0.010 &0.015 \\ \hline
27 &$(0.25,1)$ &$(-0.25,0)$ &$<0.5$ &0.007 &0.005 &0.005 &0.006 &0.009 &0.007 &0.007 &0.010 \\ \hline
28 &$(0.25,1)$ &$(-0.25,0)$ &$>0.5$ &0.007 &0.005 &0.006 &0.006 &0.009 &0.008 &0.007 &0.008 \\ \hline
29 &$(0.25,1)$ &$(0,0.25)$ &$<0.5$ &0.008 &0.006 &0.006 &0.008 &0.010 &0.010 &0.010 &0.014 \\ \hline
30 &$(0.25,1)$ &$(0,0.25)$ &$>0.5$ &0.008 &0.006 &0.006 &0.008 &0.011 &0.009 &0.009 &0.012 \\ \hline
31 &$(0.25,1)$ &$(0.25,1)$ &$<0.5$ &0.015 &0.012 &0.012 &0.018 &0.024 &0.019 &0.019 &0.029 \\ \hline
32 &$(0.25,1)$ &$(0.25,1)$ &$>0.5$ &0.015 &0.014 &0.014 &0.020 &0.027 &0.022 &0.022 &0.032 \\ \hline
\bottomrule
\end{tabular}
\end{table}

\subsection{Runtimes}\label{sec:runtimes}

In this section, we report the computational runtime of the proposed NN. The reported values correspond to inference time and were obtained on a personal computer equipped with an Intel(R) Core(TM) i7-14650HX processor (2.20 GHz) and 32 GB of RAM. 

A key advantage of deep learning–based models is their ability to perform inference in parallel without a commensurate increase in runtime. To highlight this property, we evaluate the networks on batches of 5{,}000 instances processed simultaneously. The results demonstrate that, even at this scale, the NN requires only a fraction of a second to produce predictions. In total, for 5{,}000 instances it took 0.004 seconds.

\section{Queueing network implementation} \label{sec:questa}

In this section, we show that our results can serve as a building block for a broader queueing analysis, where, combined with results from~\cite{SHERZER2025141}, we can analyze feed-forward queueing networks with superposed streams. We first provide exact details regarding the underlying queueing networks and the experiments taking place in Section~\ref{sec:queueing_disc}. Then, we present the results in Section~\ref{sec:peva}. We end with presenting inference time in Section~\ref{sec:inftimeqeusta}.

\subsection{Queueing networks discription}\label{sec:queueing_disc}

\begin{figure}
\centering
\includegraphics[scale = 0.65]{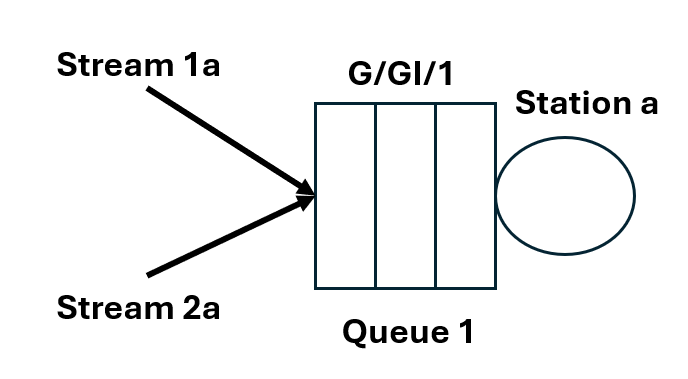}
\caption{System 1 topology.  }
\label{fig:sys1}
\end{figure}

\begin{figure}
\centering
\includegraphics[scale = 0.65]{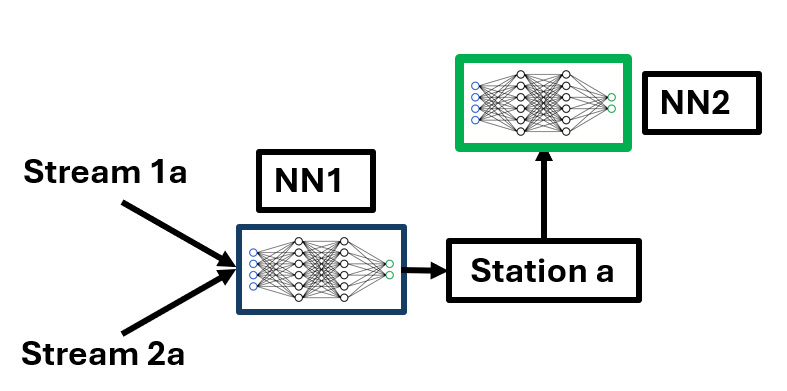}
\caption{System 1 NN workflow.  }
\label{fig:sys_1_nn}
\end{figure}

We consider two distinct queueing network topologies. The first, we denote by System 1,  consists of two non-renewal arrival processes feeding a single-server queue with renewal and general service times, as illustrated in Figure~\ref{fig:sys1}. The objective is to approximate the steady-state distribution of the number of customers in the system. To this end, we first superpose the two arrival streams and then analyze the resulting system under a $G/GI/1$ framework. 

The superposition is performed using the neural network developed in this study, denoted by \textbf{NN~1}. After merging the arrival processes, the steady-state probabilities of the corresponding $G/GI/1$ queue are approximated using the neural network introduced in~\cite{SHERZER2025141}, denoted by \textbf{NN~3}. The overall procedure is depicted in Figure~\ref{fig:sys_1_nn}.

The second topology, denoted by System 2, introduced in Section~\ref{sec:intro} and shown in Figure~\ref{fig:system2}, comprises a three-station network. Stations $a$ and $b$ each receive two non-renewal external arrival streams (Streams $(1a,2a)$ and $(1b,2b)$, respectively). At each of these stations, the two incoming streams are superposed and modeled as a $G/GI/1$ queue with renewal and general service times. The departure processes from Stations $a$ and $b$ are then routed to Station $c$, where they are again superposed. Station $c$ also operates as a $G/GI/1$ queue with renewal and general service times. The primary objective is to approximate the steady-state probabilities at Station $c$.

Figure~\ref{fig:tier2_nns} (from Section~\ref{sec:intro}) illustrates the neural networks used in this topology. For convenience, we briefly repeat the main flow. First, \textbf{NN~1} is applied to superpose Streams $1a$ and $2a$ at Station $a$, and similarly Streams $1b$ and $2b$ at Station $b$. Next, \textbf{NN~2}, developed in~\cite{SHERZER2025141}, approximates the departure processes from Stations $a$ and $b$. The resulting departure streams (namely, Streams $1c$ and $2c$) are then superposed using \textbf{NN~1}. Finally, \textbf{NN~3} is employed to approximate the steady-state probabilities at Station $c$. 

We now specify the inputs and outputs of the neural networks. For \textbf{NN~1}, the input is
\[
\Big(
log(m_{A_j}(i)),
\rho_{A_j}(k,a_1,a_2)
\Big),
\quad
j \in \{1,2\},
\quad
i \leq 5,
\quad
a_1,a_2 \leq 2,
\quad
k \leq 2,
\]
that is, the first five moments of each inter-arrival process together with the first two lag autocorrelations, including polynomial terms up to second order. The output is
\[
\Big(
log(\hat{m}_{S}(i)),
\hat{\rho}_{S}(k,a_1,a_2)
\Big),
\quad
i \leq 5,
\quad
a_1,a_2 \leq 2,
\quad
k \leq 2,
\]
corresponding to the first five moments and autocorrelation descriptors of the superposed process.

The input to \textbf{NN~2} consists of the output of \textbf{NN~1}, namely the first five moments and autocorrelation terms of the superposed non-renewal arrival process, together with the first five moments of the renewal service-time distribution:
\[
\Big(
log(\hat{m}_{S}(i)),
\hat{\rho}_{S}(k,a_1,a_2),
log(m_{\text{service}}(i))
\Big),
\quad
i \leq 5,
\quad
a_1,a_2 \leq 2,
\quad
k \leq 2,
\]
where $m_{\text{service}}(i)$ denotes the $i^{th}$ service-time moment.  Its output consists of the first five moments of the departure process and the associated autocorrelation descriptors (with $a_1,a_2 \leq 2$ and $k \leq 2$), analogous to the output structure of \textbf{NN~1}.

Finally, the input to \textbf{NN~3} is identical to that of \textbf{NN~2}. The output of \textbf{NN~2} is the steady-state probabilities where $p_i$ and $\hat{p}_i$ are the true and approximated probability of having $i$ customers in the system, respectively, truncated at $i<500$ for practical reasons.


The primary objective of this analysis is to assess the accuracy of the steady-state approximation at Station $a$ in System 1 and at Station $c$ in System 2. 
To this end, we employ two evaluation metrics. 
The first metric, originally introduced in~\cite{sherzer23}, is the Sum of Absolute Errors (SAE), which measures the absolute deviation between the predicted steady-state probabilities and the corresponding reference values. The advantage of this metric is that it is highly sensitive to any difference between actual and predicted values. It is given by

\begin{align}\label{eq:SAE}
  SAE =   \frac{1}{v}\sum_{i=1}^{v}\sum_{l=0}^{499}|p_{i}-\hat{p}_{i}|,
\end{align}   The $SAE$ is equivalent to the Wasserstein-1 measure. 

The second metric, dubbed  REM, is the Relative Error of the Mean number of customers in the system. 
\begin{align*} 
REM = 100\frac{|\sum_{i=0}^{l-1}j(p_{i}-\hat{p}_{i})|}{\sum_{i=0}^{499}j\hat{p}_{i}}.
\end{align*} 
The benefit of this measure is that it encapsulates the model's accuracy with respect to the most commonly used performance metric in queueing systems, the average number of customers in the queue. Furthermore, it enables a comparison between our ML model and existing approximation methods, since no other method aims to capture the entire distribution.

\noindent \textbf{Experiments:} We begin with a description of the experimental design for System 1. 
A total of 64 scenarios are considered, varying across the squared coefficients of variation (SCVs) of the two arrival streams and the service-time distribution, the first-lag autocorrelation of each arrival stream, and the server utilization level. 

The SCV values are partitioned into two ranges: below 3 and at least 3. 
The first-lag autocorrelation is classified as either negative or positive. 
The server utilization is divided into two regimes: below 0.7 and at least 0.7. Altogether, this results in six binary partition factors, yielding $2^6 = 64$ distinct categories.

For each category, we compute the SAE and REM. In addition, for comparison, we compute the REM by implementing the method in~\cite {WHITT201999}. Their method can both superpose and analyze the mean number of customers in the system, which makes it comparable. 

For System 2, we adopt a parallel experimental design. 
To ensure consistency between the single-station and three-station settings, the same parameter configurations are applied across all stations. 
Specifically, if the SCV of arrival Stream $1a$ at Station $a$ is below 3, then the SCV of Stream $1b$ at Station $b$ is also set below 3; the same alignment is imposed between Streams $2a$ and $2b$. 
Similarly, identical utilization regimes are enforced across the three stations.

Allowing each station to vary independently would lead to a combinatorial expansion of the scenario space, resulting in several thousand configurations and substantially reducing interpretability. 
By maintaining synchronized partitions across stations, we preserve tractability while still capturing the essential variability structure. 
As in System 1, we compare the REM results with those reported in~\cite{WHITT201999}, which is applicable to the analysis of System 2 as well.

The ground truth is obtained via discrete-event simulation. 
We implemented the simulation using the Python package \textit{SimPy}, generating $10{,}000{,}000$ arrivals for each external arrival stream. 
This corresponds to a total of $20{,}000{,}000$ and $40{,}000{,}000$ arrivals for System~1 and System~2, respectively. 
Such large sample sizes ensure highly accurate estimates of both the SAE and REM metrics.

\subsection{Performence evaluation}\label{sec:peva}

Below, we provide a performance analysis of both Systems 1 and 2.

\noindent \textbf{System 1: Performance Analysis} Table~\ref{tab:sys1} reports the steady-state performance of the proposed decomposition for System~1 across 64 structural regimes defined by variability (SCV), dependence (sign of first-lag autocorrelation), service variability, and utilization level. For each regime, we report the steady-state absolute error (SAE) of the predicted queue-length distribution, the REM of the neural-network-based prediction, and the corresponding REM obtained using Whitt’s renewal approximation.

Across all regimes, the neural decomposition maintains consistently low steady-state errors. The SAE values remain small throughout, and the associated REM values are generally modest, typically below 5\% and frequently around 2-3\%. The absence of extreme deviations suggests that the superposition operator preserves the statistical information required for accurate downstream queue-length prediction within the decomposition framework.

When both arrival streams exhibit moderate variability (SCV $<3$), prediction errors remain uniformly low across utilization levels. As arrival or service variability increases, the error gradually increases. However, even when both arrival streams and the service process have SCV $>3$, the deterioration remains moderate. This indicates that the learned mapping captures higher-order variability effects that materially influence steady-state congestion.

The regimes include combinations of negative and positive first-lag autocorrelation. No qualitative instability is observed when moving between dependence structures. While certain combinations of positive dependence and high utilization yield somewhat higher errors, the increase is smooth rather than abrupt. This behavior is consistent with the fact that short-range dependence influences congestion amplification primarily under higher traffic intensities.

We identify that higher utilization ($\rho>0.7$) leads to slightly larger deviations, reflecting the increased sensitivity of steady-state distributions in heavier traffic. Nevertheless, the magnitude of the error remains controlled across all regimes. This suggests that the decomposition remains numerically stable even in moderately congested conditions.

The final column of Table~\ref{tab:sys1} reports the performance of Whitt’s renewal approximation. In contrast to the neural decomposition, the renewal-based method exhibits substantially larger errors in most regimes, particularly when variability is high or when positive dependence is present. This difference is consistent with the structural limitation of renewal approximations, which collapse temporal dependence into low-order variability summaries.

Averaging across all 64 regimes in Table~\ref{tab:sys1}, the neural decomposition attains an average REM of 2.86\%, compared to 9.20\% for Whitt’s approximation.  This corresponds to an average relative improvement of approximately 69\%.  The advantage is particularly visible in regimes with high variability (SCV $>3$) and positive dependence, where renewal-based approximations systematically underestimate congestion amplification. In contrast, the neural model maintains stable accuracy across both variability- and dependence-driven regimes. Even at higher utilization levels ($\rho>0.7$), the increase in error remains moderate and does not exhibit structural instability. The overall average SAE is 0.03.

Overall, the results in Table~\ref{tab:sys1} indicate that the proposed decomposition provides stable and accurate steady-state predictions across heterogeneous variability, dependence, and utilization regimes. The improvement over classical approximations is most pronounced in structurally complex settings where dependence and higher-order variability materially affect congestion behavior.

\begin{table}[!htp]\centering
\caption{System 1 results}\label{tab:sys1}
\begin{tabular}{|c|c|c|c|c|c|c|c|c|c|}
\hline\hline
 & \multicolumn{3}{c|}{SCV} & \multicolumn{2}{c|}{$\rho$} &  & SAE & \multicolumn{2}{c|}{REM} \\ \hline
 & Stream 1 & Stream 2 & Service & Stream 1 & Stream 2 & Utilization & NN & NN & Whitt \\ \hline
1 &$<3$ &$<3$ &$<3$ &$<0$ &$<0$ &$<0.7$ &0.026 &2.542 &7.070 \\ \hline
2 &$<3$ &$<3$ &$<3$ &$<0$ &$<0$ &$>0.7$ &0.018 &2.882 &2.083 \\ \hline
3 &$<3$ &$<3$ &$<3$ &$<0$ &$>0$ &$<0.7$ &0.016 &0.940 &2.644 \\ \hline
4 &$<3$ &$<3$ &$<3$ &$<0$ &$>0$ &$>0.7$ &0.024 &2.661 &5.216 \\ \hline
5 &$<3$ &$<3$ &$<3$ &$>0$ &$<0$ &$<0.7$ &0.012 &2.761 &6.788 \\ \hline
6 &$<3$ &$<3$ &$<3$ &$>0$ &$<0$ &$>0.7$ &0.031 &2.813 &2.495 \\ \hline
7 &$<3$ &$<3$ &$<3$ &$>0$ &$>0$ &$<0.7$ &0.018 &1.952 &7.731 \\ \hline
8 &$<3$ &$<3$ &$<3$ &$>0$ &$>0$ &$>0.7$ &0.027 &0.946 &8.865 \\ \hline
9 &$<3$ &$<3$ &$>3$ &$<0$ &$<0$ &$<0.7$ &0.021 &1.012 &6.052 \\ \hline
10 &$<3$ &$<3$ &$>3$ &$<0$ &$<0$ &$>0.7$ &0.024 &2.258 &2.335 \\ \hline
11 &$<3$ &$<3$ &$>3$ &$<0$ &$>0$ &$<0.7$ &0.021 &2.527 &2.084 \\ \hline
12 &$<3$ &$<3$ &$>3$ &$<0$ &$>0$ &$>0.7$ &0.025 &1.077 &3.472 \\ \hline
13 &$<3$ &$<3$ &$>3$ &$>0$ &$<0$ &$<0.7$ &0.027 &2.639 &5.038 \\ \hline
14 &$<3$ &$<3$ &$>3$ &$>0$ &$<0$ &$>0.7$ &0.029 &3.977 &1.048 \\ \hline
15 &$<3$ &$<3$ &$>3$ &$>0$ &$>0$ &$<0.7$ &0.026 &3.923 &4.622 \\ \hline
16 &$<3$ &$<3$ &$>3$ &$>0$ &$>0$ &$>0.7$ &0.023 &1.997 &1.920 \\ \hline
17 &$<3$ &$>3$ &$<3$ &$<0$ &$<0$ &$<0.7$ &0.027 &3.262 &6.711 \\ \hline
18 &$<3$ &$>3$ &$<3$ &$<0$ &$<0$ &$>0.7$ &0.026 &2.782 &5.816 \\ \hline
19 &$<3$ &$>3$ &$<3$ &$<0$ &$>0$ &$<0.7$ &0.024 &2.679 &14.112 \\ \hline
20 &$<3$ &$>3$ &$<3$ &$<0$ &$>0$ &$>0.7$ &0.052 &7.640 &14.061 \\ \hline
21 &$<3$ &$>3$ &$<3$ &$>0$ &$<0$ &$<0.7$ &0.033 &7.113 &15.457 \\ \hline
22 &$<3$ &$>3$ &$<3$ &$>0$ &$<0$ &$>0.7$ &0.030 &2.700 &6.508 \\ \hline
23 &$<3$ &$>3$ &$<3$ &$>0$ &$>0$ &$<0.7$ &0.025 &4.189 &15.650 \\ \hline
24 &$<3$ &$>3$ &$<3$ &$>0$ &$>0$ &$>0.7$ &0.017 &2.417 &8.424 \\ \hline
25 &$<3$ &$>3$ &$>3$ &$<0$ &$<0$ &$<0.7$ &0.040 &2.216 &9.471 \\ \hline
26 &$<3$ &$>3$ &$>3$ &$<0$ &$<0$ &$>0.7$ &0.046 &1.917 &5.072 \\ \hline
27 &$<3$ &$>3$ &$>3$ &$<0$ &$>0$ &$<0.7$ &0.049 &3.990 &13.567 \\ \hline
28 &$<3$ &$>3$ &$>3$ &$<0$ &$>0$ &$>0.7$ &0.042 &2.155 &6.428 \\ \hline
29 &$<3$ &$>3$ &$>3$ &$>0$ &$<0$ &$<0.7$ &0.044 &4.444 &14.270 \\ \hline
30 &$<3$ &$>3$ &$>3$ &$>0$ &$<0$ &$>0.7$ &0.049 &1.840 &5.427 \\ \hline
31 &$<3$ &$>3$ &$>3$ &$>0$ &$>0$ &$<0.7$ &0.033 &4.015 &12.979 \\ \hline
32 &$<3$ &$>3$ &$>3$ &$>0$ &$>0$ &$>0.7$ &0.049 &3.002 &8.178 \\ \hline
33 &$>3$ &$<3$ &$<3$ &$<0$ &$<0$ &$<0.7$ &0.018 &2.810 &10.705 \\ \hline
34 &$>3$ &$<3$ &$<3$ &$<0$ &$<0$ &$>0.7$ &0.020 &1.246 &5.179 \\ \hline
35 &$>3$ &$<3$ &$<3$ &$<0$ &$>0$ &$<0.7$ &0.023 &5.163 &15.201 \\ \hline
36 &$>3$ &$<3$ &$<3$ &$<0$ &$>0$ &$>0.7$ &0.022 &2.334 &7.146 \\ \hline
37 &$>3$ &$<3$ &$<3$ &$>0$ &$<0$ &$<0.7$ &0.031 &4.394 &15.301 \\ \hline
38 &$>3$ &$<3$ &$<3$ &$>0$ &$<0$ &$>0.7$ &0.024 &3.077 &8.810 \\ \hline
39 &$>3$ &$<3$ &$<3$ &$>0$ &$>0$ &$<0.7$ &0.026 &4.251 &14.599 \\ \hline
40 &$>3$ &$<3$ &$<3$ &$>0$ &$>0$ &$>0.7$ &0.026 &3.759 &9.450 \\ \hline
41 &$>3$ &$<3$ &$>3$ &$<0$ &$<0$ &$<0.7$ &0.018 &2.297 &9.594 \\ \hline
42 &$>3$ &$<3$ &$>3$ &$<0$ &$<0$ &$>0.7$ &0.045 &3.411 &5.674 \\ \hline
43 &$>3$ &$<3$ &$>3$ &$<0$ &$>0$ &$<0.7$ &0.049 &0.044 &5.886 \\ \hline
44 &$>3$ &$<3$ &$>3$ &$<0$ &$>0$ &$>0.7$ &0.051 &1.643 &5.824 \\ \hline
45 &$>3$ &$<3$ &$>3$ &$>0$ &$<0$ &$<0.7$ &0.043 &3.530 &14.329 \\ \hline
46 &$>3$ &$<3$ &$>3$ &$>0$ &$<0$ &$>0.7$ &0.046 &2.300 &6.774 \\ \hline
47 &$>3$ &$<3$ &$>3$ &$>0$ &$>0$ &$<0.7$ &0.045 &1.833 &12.684 \\ \hline
48 &$>3$ &$<3$ &$>3$ &$>0$ &$>0$ &$>0.7$ &0.050 &2.890 &8.384 \\ \hline
49 &$>3$ &$>3$ &$<3$ &$<0$ &$<0$ &$<0.7$ &0.026 &5.701 &18.022 \\ \hline
50 &$>3$ &$>3$ &$<3$ &$<0$ &$<0$ &$>0.7$ &0.022 &2.238 &7.114 \\ \hline
51 &$>3$ &$>3$ &$<3$ &$<0$ &$>0$ &$<0.7$ &0.012 &2.075 &17.823 \\ \hline
52 &$>3$ &$>3$ &$<3$ &$<0$ &$>0$ &$>0.7$ &0.021 &3.595 &5.130 \\ \hline
53 &$>3$ &$>3$ &$<3$ &$>0$ &$<0$ &$<0.7$ &0.022 &4.312 &19.844 \\ \hline
54 &$>3$ &$>3$ &$<3$ &$>0$ &$<0$ &$>0.7$ &0.031 &3.072 &5.975 \\ \hline
55 &$>3$ &$>3$ &$<3$ &$>0$ &$>0$ &$<0.7$ &0.012 &2.752 &27.725 \\ \hline
56 &$>3$ &$>3$ &$<3$ &$>0$ &$>0$ &$>0.7$ &0.028 &1.045 &7.694 \\ \hline
57 &$>3$ &$>3$ &$>3$ &$<0$ &$<0$ &$<0.7$ &0.031 &2.588 &12.053 \\ \hline
58 &$>3$ &$>3$ &$>3$ &$<0$ &$<0$ &$>0.7$ &0.036 &0.916 &4.175 \\ \hline
59 &$>3$ &$>3$ &$>3$ &$<0$ &$>0$ &$<0.7$ &0.034 &1.701 &14.383 \\ \hline
60 &$>3$ &$>3$ &$>3$ &$<0$ &$>0$ &$>0.7$ &0.046 &2.545 &3.302 \\ \hline
61 &$>3$ &$>3$ &$>3$ &$>0$ &$<0$ &$<0.7$ &0.033 &2.999 &14.275 \\ \hline
62 &$>3$ &$>3$ &$>3$ &$>0$ &$<0$ &$>0.7$ &0.041 &0.734 &8.348 \\ \hline
63 &$>3$ &$>3$ &$>3$ &$>0$ &$>0$ &$<0.7$ &0.041 &4.927 &22.213 \\ \hline
64 &$>3$ &$>3$ &$>3$ &$>0$ &$>0$ &$>0.7$ &0.047 &3.567 &13.549 \\ \hline\hline
\end{tabular}
\end{table}
\textbf{System 2: Performance Analysis} Table~\ref{tab:sys2} reports the steady-state performance at Station $c$ in System~2 across the same 64 structural regimes used for System~1. Across all regimes, the SAE values remain controlled, indicating that the composed pipeline
(superposition at Stations $a$ and $b$, departure approximation from $a$ and $b$, followed by superposition at $c$ and steady-state inference at $c$) does not induce a qualitative loss of distributional fidelity.
As expected, SAE tends to increase when the variability of the inputs increases and when the system operates closer to heavy traffic, reflecting the sensitivity of stationary distributions under congestion and the accumulation of approximation errors along the network path. Nevertheless, no regime exhibits a structural breakdown, suggesting numerical stability of the superposition departure composition at the network level.

Table~\ref{tab:sys2} also enables a direct mean-performance comparison against~\cite{WHITT201999} through the REM metric.
Overall, the proposed neural decomposition achieves substantially smaller REM values than the Whitt benchmark in the large majority of regimes, averaging 4.177\% in our method against 10.733\% in Whitt.   The improvement is particularly pronounced in structurally complex settings (high variability and/or positive dependence) where higher-order variability and short-range dependence materially affect congestion and where classic surrogate models may lose critical information.
Importantly, the relative behavior across regimes is consistent with queueing intuition: positive dependence combined with high variability yields the largest deviations, and high utilization amplifies sensitivity because small descriptor errors translate into larger differences in the expected number of customers in the system. The overall average SAE is 0.07.

Relative to Table~\ref{tab:sys1}, the overall error levels in System~2 are slightly higher, which is expected due to error propagation across multiple neural components and the additional interaction created by merging two approximated departure streams at Station $c$.
Crucially, however, the qualitative patterns remain the same: (i) low-variability regimes yield uniformly small errors, (ii) high SCV and positive dependence increase deviations gradually, and (iii) heavy traffic amplifies sensitivity without inducing instability. 

In summary, the results in Table~\ref{tab:sys2} demonstrate that the proposed superposition departure steady-state decomposition remains accurate and robust in a three-station feed-forward network while also improving mean-performance accuracy relative to an established benchmark.


\begin{table}[!htp]\centering
\caption{System 2 results}\label{tab:sys2}
\begin{tabular}{|l|r|r|r|r|r|r|r|r|r|r|}\hline
&\multicolumn{3}{c|}{SCV } &\multicolumn{2}{c|}{$\rho$ } & & &\multicolumn{2}{c|}{REM} \\ \hline
&Stream 1 & Stream 2 & Service &Stream 1 &Stream 2 &utilization &SAE &NN &Whitt \\ \hline
1 &$<3$ &$<3$ &$<3$ &$<0$ &$<0$ &$<0.7$ &0.064 &3.132 &8.734 \\ \hline
2 &$<3$ &$<3$ &$<3$ &$<0$ &$<0$ &$>0.7$ &0.086 &2.615 &2.652 \\ \hline
3 &$<3$ &$<3$ &$<3$ &$<0$ &$>0$ &$<0.7$ &0.074 &4.310 &4.332 \\ \hline
4 &$<3$ &$<3$ &$<3$ &$<0$ &$>0$ &$>0.7$ &0.078 &4.280 &7.753 \\ \hline
5 &$<3$ &$<3$ &$<3$ &$>0$ &$<0$ &$<0.7$ &0.073 &4.129 &7.199 \\ \hline
6 &$<3$ &$<3$ &$<3$ &$>0$ &$<0$ &$>0.7$ &0.061 &2.583 &4.272 \\ \hline
7 &$<3$ &$<3$ &$<3$ &$>0$ &$>0$ &$<0.7$ &0.068 &4.060 &8.950 \\ \hline
8 &$<3$ &$<3$ &$<3$ &$>0$ &$>0$ &$>0.7$ &0.076 &4.978 &10.426 \\ \hline
9 &$<3$ &$<3$ &$>3$ &$<0$ &$<0$ &$<0.7$ &0.100 &4.944 &7.878 \\ \hline
10 &$<3$ &$<3$ &$>3$ &$<0$ &$<0$ &$>0.7$ &0.098 &5.042 &3.670 \\ \hline
11 &$<3$ &$<3$ &$>3$ &$<0$ &$>0$ &$<0.7$ &0.019 &2.416 &2.889 \\ \hline
12 &$<3$ &$<3$ &$>3$ &$<0$ &$>0$ &$>0.7$ &0.068 &5.236 &4.034 \\ \hline
13 &$<3$ &$<3$ &$>3$ &$>0$ &$<0$ &$<0.7$ &0.105 &4.697 &8.017 \\ \hline
14 &$<3$ &$<3$ &$>3$ &$>0$ &$<0$ &$>0.7$ &0.080 &6.470 &1.183 \\ \hline
15 &$<3$ &$<3$ &$>3$ &$>0$ &$>0$ &$<0.7$ &0.100 &8.079 &5.407 \\ \hline
16 &$<3$ &$<3$ &$>3$ &$>0$ &$>0$ &$>0.7$ &0.071 &2.880 &3.887 \\ \hline
17 &$<3$ &$>3$ &$<3$ &$<0$ &$<0$ &$<0.7$ &0.057 &3.817 &7.758 \\ \hline
18 &$<3$ &$>3$ &$<3$ &$<0$ &$<0$ &$>0.7$ &0.083 &2.774 &7.607 \\ \hline
19 &$<3$ &$>3$ &$<3$ &$<0$ &$>0$ &$<0.7$ &0.047 &3.362 &14.232 \\ \hline
20 &$<3$ &$>3$ &$<3$ &$<0$ &$>0$ &$>0.7$ &0.088 &8.680 &16.428 \\ \hline
21 &$<3$ &$>3$ &$<3$ &$>0$ &$<0$ &$<0.7$ &0.062 &2.508 &17.973 \\ \hline
22 &$<3$ &$>3$ &$<3$ &$>0$ &$<0$ &$>0.7$ &0.052 &1.827 &8.656 \\ \hline
23 &$<3$ &$>3$ &$<3$ &$>0$ &$>0$ &$<0.7$ &0.092 &4.825 &16.064 \\ \hline
24 &$<3$ &$>3$ &$<3$ &$>0$ &$>0$ &$>0.7$ &0.066 &5.647 &10.220 \\ \hline
25 &$<3$ &$>3$ &$>3$ &$<0$ &$<0$ &$<0.7$ &0.059 &3.100 &10.596 \\ \hline
26 &$<3$ &$>3$ &$>3$ &$<0$ &$<0$ &$>0.7$ &0.062 &4.039 &5.305 \\ \hline
27 &$<3$ &$>3$ &$>3$ &$<0$ &$>0$ &$<0.7$ &0.055 &7.941 &15.212 \\ \hline
28 &$<3$ &$>3$ &$>3$ &$<0$ &$>0$ &$>0.7$ &0.109 &4.923 &8.442 \\ \hline
29 &$<3$ &$>3$ &$>3$ &$>0$ &$<0$ &$<0.7$ &0.069 &3.609 &17.006 \\ \hline
30 &$<3$ &$>3$ &$>3$ &$>0$ &$<0$ &$>0.7$ &0.097 &5.436 &8.411 \\ \hline
31 &$<3$ &$>3$ &$>3$ &$>0$ &$>0$ &$<0.7$ &0.071 &5.989 &13.079 \\ \hline
32 &$<3$ &$>3$ &$>3$ &$>0$ &$>0$ &$>0.7$ &0.060 &5.519 &9.245 \\ \hline
33 &$>3$ &$<3$ &$<3$ &$<0$ &$<0$ &$<0.7$ &0.041 &3.156 &11.463 \\ \hline
34 &$>3$ &$<3$ &$<3$ &$<0$ &$<0$ &$>0.7$ &0.065 &7.980 &5.924 \\ \hline
35 &$>3$ &$<3$ &$<3$ &$<0$ &$>0$ &$<0.7$ &0.060 &3.318 &16.717 \\ \hline
36 &$>3$ &$<3$ &$<3$ &$<0$ &$>0$ &$>0.7$ &0.089 &5.818 &8.150 \\ \hline
37 &$>3$ &$<3$ &$<3$ &$>0$ &$<0$ &$<0.7$ &0.088 &4.034 &16.438 \\ \hline
38 &$>3$ &$<3$ &$<3$ &$>0$ &$<0$ &$>0.7$ &0.132 &6.459 &10.588 \\ \hline
39 &$>3$ &$<3$ &$<3$ &$>0$ &$>0$ &$<0.7$ &0.056 &3.329 &15.967 \\ \hline
40 &$>3$ &$<3$ &$<3$ &$>0$ &$>0$ &$>0.7$ &0.105 &5.483 &10.723 \\ \hline
41 &$>3$ &$<3$ &$>3$ &$<0$ &$<0$ &$<0.7$ &0.074 &4.566 &11.647 \\ \hline
42 &$>3$ &$<3$ &$>3$ &$<0$ &$<0$ &$>0.7$ &0.088 &5.192 &8.578 \\ \hline
43 &$>3$ &$<3$ &$>3$ &$<0$ &$>0$ &$<0.7$ &0.084 &4.350 &8.289 \\ \hline
44 &$>3$ &$<3$ &$>3$ &$<0$ &$>0$ &$>0.7$ &0.130 &3.642 &7.952 \\ \hline
45 &$>3$ &$<3$ &$>3$ &$>0$ &$<0$ &$<0.7$ &0.095 &3.710 &14.836 \\ \hline
46 &$>3$ &$<3$ &$>3$ &$>0$ &$<0$ &$>0.7$ &0.089 &4.901 &8.611 \\ \hline
47 &$>3$ &$<3$ &$>3$ &$>0$ &$>0$ &$<0.7$ &0.080 &2.169 &14.809 \\ \hline
48 &$>3$ &$<3$ &$>3$ &$>0$ &$>0$ &$>0.7$ &0.083 &4.747 &9.626 \\ \hline
49 &$>3$ &$>3$ &$<3$ &$<0$ &$<0$ &$<0.7$ &0.069 &2.689 &19.860 \\ \hline
50 &$>3$ &$>3$ &$<3$ &$<0$ &$<0$ &$>0.7$ &0.068 &1.607 &9.912 \\ \hline
51 &$>3$ &$>3$ &$<3$ &$<0$ &$>0$ &$<0.7$ &0.031 &2.917 &20.637 \\ \hline
52 &$>3$ &$>3$ &$<3$ &$<0$ &$>0$ &$>0.7$ &0.021 &5.298 &5.801 \\ \hline
53 &$>3$ &$>3$ &$<3$ &$>0$ &$<0$ &$<0.7$ &0.082 &2.702 &22.769 \\ \hline
54 &$>3$ &$>3$ &$<3$ &$>0$ &$<0$ &$>0.7$ &0.106 &1.125 &6.621 \\ \hline
55 &$>3$ &$>3$ &$<3$ &$>0$ &$>0$ &$<0.7$ &0.054 &3.779 &30.688 \\ \hline
56 &$>3$ &$>3$ &$<3$ &$>0$ &$>0$ &$>0.7$ &0.062 &4.482 &8.075 \\ \hline
57 &$>3$ &$>3$ &$>3$ &$<0$ &$<0$ &$<0.7$ &0.085 &0.893 &14.404 \\ \hline
58 &$>3$ &$>3$ &$>3$ &$<0$ &$<0$ &$>0.7$ &0.095 &2.166 &7.169 \\ \hline
59 &$>3$ &$>3$ &$>3$ &$<0$ &$>0$ &$<0.7$ &0.086 &3.308 &16.927 \\ \hline
60 &$>3$ &$>3$ &$>3$ &$<0$ &$>0$ &$>0.7$ &0.073 &1.021 &4.027 \\ \hline
61 &$>3$ &$>3$ &$>3$ &$>0$ &$<0$ &$<0.7$ &0.084 &4.678 &14.520 \\ \hline
62 &$>3$ &$>3$ &$>3$ &$>0$ &$<0$ &$>0.7$ &0.082 &3.272 &8.944 \\ \hline
63 &$>3$ &$>3$ &$>3$ &$>0$ &$>0$ &$<0.7$ &0.082 &4.662 &23.748 \\ \hline
64 &$>3$ &$>3$ &$>3$ &$>0$ &$>0$ &$>0.7$ &0.072 &6.008 &14.985 \\ \hline
\end{tabular}
\end{table}
\subsection{Inference time}\label{sec:inftimeqeusta}

The overall inference time remains extremely small, even though the procedure involves multiple stages, and it can be executed in parallel over a large number of instances. In this study, we evaluated 64 queueing configurations for System~1 and 64 for System~2. Scaling the framework to a substantially larger number of configurations would have a negligible impact on the total inference time. The measured runtimes were 0.015 and 0.05 seconds for Systems~1 and~2, respectively.

The reported inference time accounts not only for the forward pass through the neural networks but also for the auxiliary computations required by the framework. In particular, each network requires a preprocessing step that normalizes time so that the mean inter-arrival equals one. For \textbf{NN~1}, as described in Section~\ref{sec:data_generation}, the MAP with the largest mean is scaled to have a unit mean. Similarly, for \textbf{NN~2} and \textbf{NN~3}, the arrival process is normalized to have a mean of 1. Therefore, prior to applying each network, an appropriate time-scaling transformation is performed.

\begin{figure}[ht]
    \centering
    \begin{subfigure}[b]{0.49\textwidth}
        \centering
        \includegraphics[width=\textwidth]{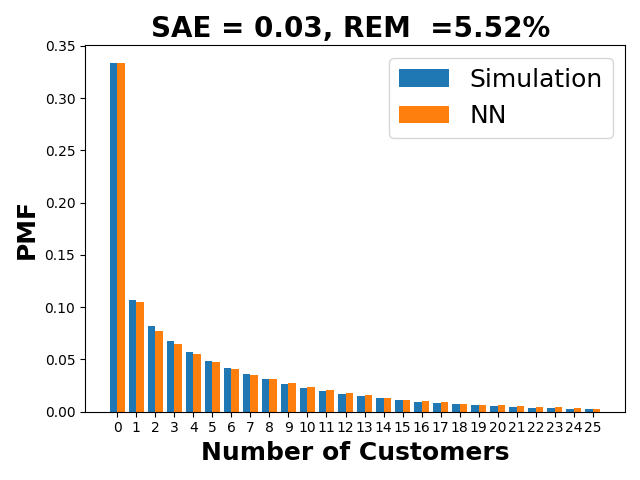}
        \caption{Example of System 1. }
\label{fig:exmaple_sys1}
    \end{subfigure}
    \hfill
    \begin{subfigure}[b]{0.49\textwidth}
        \centering
        \includegraphics[width=\textwidth]{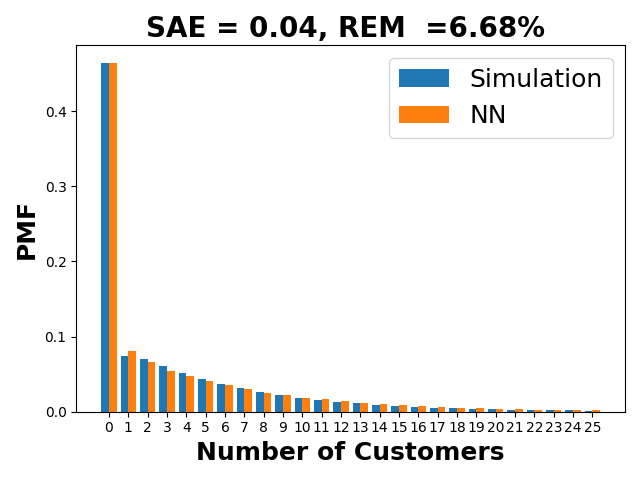}
        \caption{Example of System 2.}
\label{fig:exmaple_sys2}
    \end{subfigure}
    \caption{Examples of steady-state probabilities approximations}
    \label{fig:probs_examples}
\end{figure}

For illustrative purposes, we present representative examples comparing the predicted steady-state probabilities with the ground truth obtained via simulation. In both cases, the selected instances exhibit average performance (slightly worse in terms of REM), thereby demonstrating the quality of fit for a typical scenario. 

\section{Discussion and Conclusions}\label{sec:conclu}

\begin{figure}
\centering
\includegraphics[scale = 0.3]{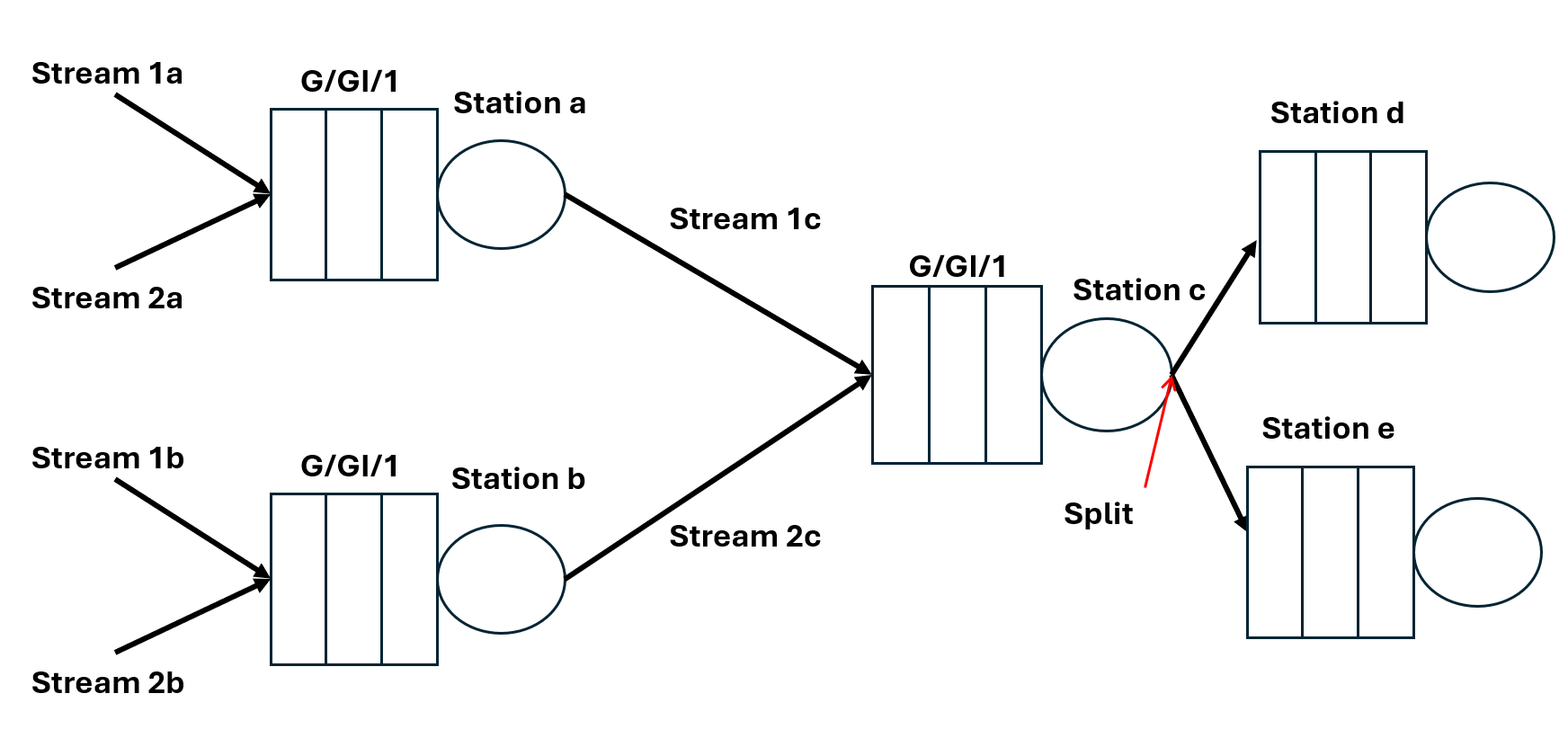}
\caption{Queueing networks with splitting streams.  }
\label{fig:sys3_split}
\end{figure}

The results presented in this paper demonstrate that the proposed neural superposition operator can function as a foundational component for scalable queueing network analysis. When integrated with the steady-state and departure-process neural modules developed in~\cite{SHERZER2025141}, the framework facilitates decomposition-based evaluation of feed-forward queueing systems with merging flows. We next introduce a natural yet practically important extension, discuss the methodological limitations, and highlight the key advantages of the approach.

\subsection{Extension to General Feed-Forward Networks}

The extension of the proposed framework to a broader class of feed-forward networks relies on introducing an additional neural module that performs \emph{flow splitting} (also referred to as thinning). Specifically, beyond the superposition operator developed in this paper and the departure and steady-state modules introduced in~\cite{SHERZER2025141}, we propose a fourth neural network that approximates the statistical descriptors of the processes generated when a non-renewal arrival stream is split into two or more downstream streams.

As illustrated in Figure~\ref{fig:sys3_split}, a generic feed-forward architecture may include both merging and splitting nodes. At merging nodes, the superposition operator computes the descriptors of the aggregated stream. At service nodes, the departure operator characterizes the output process and  approximates the steady-state prbabilities. To complete the building-block structure required for general feed-forward networks, a splitting operator would take as input the moments and autocorrelation descriptors of a single non-renewal process and output the corresponding descriptors of the resulting downstream processes after probabilistic routing.

With these four components-superposition, departure, steady-state evaluation, and splitting-one can recursively analyze arbitrary feed-forward queueing networks. The network is first topologically ordered. At each node, incoming streams are merged using the superposition network, processed through the service node via the departure and steady-state networks, and then distributed downstream through the splitting network. Since renewal processes are a special case of non-renewal inputs, this framework remains fully general.

In this manner, the methodology is not restricted to tandem or simple merging structures. Instead, it naturally extends to multi-layer feed-forward systems with arbitrary combinations of merging and splitting points, without constructing the full multidimensional state space.

\subsection{Error Propagation and Domain Specialization}

A fundamental limitation of any decomposition-based approximation framework is the accumulation of approximation errors along the network. In the proposed approach, arrival descriptors are propagated downstream: the output of one neural module becomes the input of the next. Consequently, small inaccuracies in moment or autocorrelation prediction may compound as the network depth increases.

The experimental results indicate that, for the topologies examined in Section~\ref{sec:questa}, the accumulated error remains controlled. However, for deeper networks or extreme variability regimes, error amplification may become non-negligible. This highlights the importance of training highly accurate single-node modules.

One possible strategy to mitigate propagation effects is domain specialization. Instead of training a single universal model across an extremely broad variability domain, one may construct specialized networks tailored to specific SCV and correlation ranges. Although this reduces generality, it may significantly improve local accuracy and reduce cumulative errors in large-scale systems. Such trade-offs between universality and precision constitute an important direction for future research.

Another promising direction is the incorporation of uncertainty quantification mechanisms, enabling downstream nodes to account for upstream prediction variability. This could potentially stabilize recursive evaluations in deeper networks.

\subsection{Computational Scalability}

A key advantage of the proposed framework is its computational efficiency. Exact MAP superposition leads to multiplicative state-space growth, rendering classical methods infeasible for large networks. In contrast, the neural superposition operator operates purely on low-dimensional statistical descriptors. Inference is nearly instantaneous and can be performed in parallel over thousands of instances. 

This computational scalability makes the framework particularly attractive for:
\begin{itemize}
    \item real-time performance evaluation,
    \item large-scale design-space exploration,
    \item simulation acceleration,
    \item integration within optimization loops.
\end{itemize}

\subsection{Conclusion}

This paper introduced a neural-network-based superposition operator for merging non-renewal arrival streams. The operator maps low-order moments and short-range autocorrelation descriptors of individual streams to the corresponding descriptors of their superposition. Extensive experiments demonstrate that the learned mapping is identifiable from a compact statistical representation and significantly outperforms classical renewal-based approximations.

When combined with previously developed neural modules for departure-process characterization and steady-state distribution prediction, the proposed operator enables scalable analysis of general feed-forward single-server queueing networks with merging flows. The framework avoids renewal assumptions, bypasses MAP state-space explosion, and preserves higher-order variability and dependence information essential for accurate distributional predictions.

Overall, the present work advances a modular, data-driven paradigm for queueing network analysis. By replacing intractable stochastic constructions with learned statistical operators, it opens a path toward tractable yet statistically rich modeling of complex non-Markovian networks. Future research directions include extending the framework to networks with feedback loops, incorporating uncertainty quantification, and integrating the methodology within optimization and control settings.

\setcitestyle{authoryear, maxnames=1}



\backmatter








\section*{Acknowledgments}
This research was partially supported by the Israel Science Foundation (ISF) Grant No. 1587/25 awarded to Eliran Sherzer.


\begin{thebibliography}{}
\renewcommand{\doi}[1]{\url{https://doi.org/#1}}
\bibcommenthead

\bibitem [\protect \citeauthoryear {%
Albin%
}{%
Albin%
}{%
{\protect \APACyear {1984}}%
}]{%
Albin1984}
\APACinsertmetastar {%
Albin1984}%
\begin{APACrefauthors}%
Albin, S.L.%
\end{APACrefauthors}%
\unskip\
\newblock
\APACrefYearMonthDay{1984}{}{}.
\newblock
{\BBOQ}\APACrefatitle {Approximating a Point Process by a Renewal Process: Superposition Arrival Processes to Queues} {Approximating a point process by a renewal process: Superposition arrival processes to queues}.{\BBCQ}
\newblock
\APACjournalVolNumPages{Operations Research}{32}{5}{1133--1162,}
\newblock

\newblock

\PrintBackRefs{\CurrentBib}

\bibitem [\protect \citeauthoryear {%
Asmussen%
}{%
Asmussen%
}{%
{\protect \APACyear {2003}}%
}]{%
Asmussen2003}
\APACinsertmetastar {%
Asmussen2003}%
\begin{APACrefauthors}%
Asmussen, S.%
\end{APACrefauthors}%
\unskip\
\newblock
\APACrefYear{2003}.
\newblock
\APACrefbtitle {Applied Probability and Queues} {Applied probability and queues}\ (\PrintOrdinal{2nd}\ \BEd).
\newblock
\APACaddressPublisher{}{Springer}.
\PrintBackRefs{\CurrentBib}

\bibitem [\protect \citeauthoryear {%
Bandi%
\ \protect \BOthers {.}}{%
Bandi%
\ \protect \BOthers {.}}{%
{\protect \APACyear {2015}}%
}]{%
doi:10.1287/opre.2015.1367}
\APACinsertmetastar {%
doi:10.1287/opre.2015.1367}%
\begin{APACrefauthors}%
Bandi, C.%
, Bertsimas, D.%
\BCBL {} Youssef, N.%
\end{APACrefauthors}%
\unskip\
\newblock
\APACrefYearMonthDay{2015}{}{}.
\newblock
{\BBOQ}\APACrefatitle {Robust Queueing Theory} {Robust queueing theory}.{\BBCQ}
\newblock
\APACjournalVolNumPages{Operations Research}{63}{3}{676-700,}
\newblock
\begin{APACrefDOI} \doi{10.1287/opre.2015.1367} \end{APACrefDOI}
\newblock
\begin{APACrefURL} {https://doi.org/10.1287/opre.2015.1367} \end{APACrefURL}
\newblock
{\href{https://arxiv.org/abs/https://doi.org/10.1287/opre.2015.1367}{{https://doi.org/10.1287/opre.2015.1367}}}
\newblock

\PrintBackRefs{\CurrentBib}

\bibitem [\protect \citeauthoryear {%
Baron%
\ \protect \BOthers {.}}{%
Baron%
\ \protect \BOthers {.}}{%
{\protect \APACyear {2024}}%
}]{%
sherzer23}
\APACinsertmetastar {%
sherzer23}%
\begin{APACrefauthors}%
Baron, O.%
, Krass, D.%
, Senderovich, A.%
\BCBL {} Sherzer, E.%
\end{APACrefauthors}%
\unskip\
\newblock
\APACrefYearMonthDay{2024}{}{}.
\newblock
{\BBOQ}\APACrefatitle {Supervised ML for Solving the GI/GI/1 Queue} {Supervised ml for solving the gi/gi/1 queue}.{\BBCQ}
\newblock
\APACjournalVolNumPages{INFORMS Journal on Computing}{36}{3}{766-786,}
\newblock
\begin{APACrefDOI} \doi{10.1287/ijoc.2022.0263} \end{APACrefDOI}
\newblock

\newblock

\PrintBackRefs{\CurrentBib}

\bibitem [\protect \citeauthoryear {%
Chung%
\ \protect \BOthers {.}}{%
Chung%
\ \protect \BOthers {.}}{%
{\protect \APACyear {2015}}%
}]{%
NIPS2015_b618c321}
\APACinsertmetastar {%
NIPS2015_b618c321}%
\begin{APACrefauthors}%
Chung, J.%
, Kastner, K.%
, Dinh, L.%
, Goel, K.%
, Courville, A.C.%
\BCBL {} Bengio, Y.%
\end{APACrefauthors}%
\unskip\
\newblock
\APACrefYearMonthDay{2015}{}{}.
\newblock
{\BBOQ}\APACrefatitle {A Recurrent Latent Variable Model for Sequential Data} {A recurrent latent variable model for sequential data}.{\BBCQ}
\newblock
 C.~Cortes, N.~Lawrence, D.~Lee, M.~Sugiyama\BCBL {}\ \BBA {} R.~Garnett\ (\BEDS), \APACrefbtitle {Advances in Neural Information Processing Systems} {Advances in neural information processing systems}\ (\BVOL~28).
\newblock
\APACaddressPublisher{}{Curran Associates, Inc.}
\PrintBackRefs{\CurrentBib}

\bibitem [\protect \citeauthoryear {%
Glynn%
\ \BBA {} Whitt%
}{%
Glynn%
\ \BBA {} Whitt%
}{%
{\protect \APACyear {1994}}%
}]{%
GlynnWhitt1994}
\APACinsertmetastar {%
GlynnWhitt1994}%
\begin{APACrefauthors}%
Glynn, P.W.%
\BCBT {}\ \BBA {} Whitt, W.%
\end{APACrefauthors}%
\unskip\
\newblock
\APACrefYearMonthDay{1994}{}{}.
\newblock
{\BBOQ}\APACrefatitle {Logarithmic Asymptotics for Steady-State Tail Probabilities in a Single-Server Queue} {Logarithmic asymptotics for steady-state tail probabilities in a single-server queue}.{\BBCQ}
\newblock
\APACjournalVolNumPages{Journal of Applied Probability}{31}{}{131--156,}
\newblock

\newblock

\PrintBackRefs{\CurrentBib}

\bibitem [\protect \citeauthoryear {%
He%
\ \BBA {} Neuts%
}{%
He%
\ \BBA {} Neuts%
}{%
{\protect \APACyear {1998}}%
}]{%
HeNeuts1998}
\APACinsertmetastar {%
HeNeuts1998}%
\begin{APACrefauthors}%
He, Q.%
\BCBT {}\ \BBA {} Neuts, M.F.%
\end{APACrefauthors}%
\unskip\
\newblock
\APACrefYearMonthDay{1998}{}{}.
\newblock
{\BBOQ}\APACrefatitle {Markov Chains with Marked Transitions} {Markov chains with marked transitions}.{\BBCQ}
\newblock
\APACjournalVolNumPages{Probability in the Engineering and Informational Sciences}{12}{1}{51--77,}
\newblock

\newblock

\PrintBackRefs{\CurrentBib}

\bibitem [\protect \citeauthoryear {%
Horv{\'a}th%
\ \BBA {} Telek%
}{%
Horv{\'a}th%
\ \BBA {} Telek%
}{%
{\protect \APACyear {2002}}%
}]{%
HorvathTelek2002}
\APACinsertmetastar {%
HorvathTelek2002}%
\begin{APACrefauthors}%
Horv{\'a}th, G.%
\BCBT {}\ \BBA {} Telek, M.%
\end{APACrefauthors}%
\unskip\
\newblock
\APACrefYearMonthDay{2002}{}{}.
\newblock
{\BBOQ}\APACrefatitle {Matching Marginal Moments and Lag Correlations of Markovian Arrival Processes} {Matching marginal moments and lag correlations of markovian arrival processes}.{\BBCQ}
\newblock
\APACjournalVolNumPages{Performance Evaluation}{48}{1--4}{193--209,}
\newblock

\newblock

\PrintBackRefs{\CurrentBib}

\bibitem [\protect \citeauthoryear {%
Lucantoni%
\ \BBA {} Meier-Hellstern%
}{%
Lucantoni%
\ \BBA {} Meier-Hellstern%
}{%
{\protect \APACyear {1990}}%
}]{%
Lucantoni1990}
\APACinsertmetastar {%
Lucantoni1990}%
\begin{APACrefauthors}%
Lucantoni, D.M.%
\BCBT {}\ \BBA {} Meier-Hellstern, K.S.%
\end{APACrefauthors}%
\unskip\
\newblock
\APACrefYearMonthDay{1990}{}{}.
\newblock
{\BBOQ}\APACrefatitle {A Single-Server Queue with Server Vacations and a Class of Non-Renewal Arrival Processes} {A single-server queue with server vacations and a class of non-renewal arrival processes}.{\BBCQ}
\newblock
\APACjournalVolNumPages{Advances in Applied Probability}{22}{3}{676--705,}
\newblock

\newblock

\PrintBackRefs{\CurrentBib}

\bibitem [\protect \citeauthoryear {%
Neuts%
}{%
Neuts%
}{%
{\protect \APACyear {1979}}%
}]{%
Neuts1979}
\APACinsertmetastar {%
Neuts1979}%
\begin{APACrefauthors}%
Neuts, M.F.%
\end{APACrefauthors}%
\unskip\
\newblock
\APACrefYear{1979}.
\newblock
\APACrefbtitle {Matrix-Geometric Solutions in Stochastic Models} {Matrix-geometric solutions in stochastic models}.
\newblock
\APACaddressPublisher{}{Johns Hopkins University Press}.
\PrintBackRefs{\CurrentBib}

\bibitem [\protect \citeauthoryear {%
Newell%
}{%
Newell%
}{%
{\protect \APACyear {1984}}%
}]{%
Newell1984}
\APACinsertmetastar {%
Newell1984}%
\begin{APACrefauthors}%
Newell, G.F.%
\end{APACrefauthors}%
\unskip\
\newblock
\APACrefYearMonthDay{1984}{}{}.
\newblock
{\BBOQ}\APACrefatitle {Approximations for Superposition Arrival Processes in Queues} {Approximations for superposition arrival processes in queues}.{\BBCQ}
\newblock
\APACjournalVolNumPages{Management Science}{30}{5}{623--632,}
\newblock

\newblock

\PrintBackRefs{\CurrentBib}

\bibitem [\protect \citeauthoryear {%
Reiman%
}{%
Reiman%
}{%
{\protect \APACyear {1984}}%
}]{%
Reiman1984}
\APACinsertmetastar {%
Reiman1984}%
\begin{APACrefauthors}%
Reiman, M.I.%
\end{APACrefauthors}%
\unskip\
\newblock
\APACrefYearMonthDay{1984}{}{}.
\newblock
{\BBOQ}\APACrefatitle {Open Queueing Networks in Heavy Traffic} {Open queueing networks in heavy traffic}.{\BBCQ}
\newblock
\APACjournalVolNumPages{Mathematics of Operations Research}{9}{3}{441-458,}
\newblock
\begin{APACrefDOI} \doi{10.1287/moor.9.3.441} \end{APACrefDOI}
\newblock
\begin{APACrefURL} {https://doi.org/10.1287/moor.9.3.441} \end{APACrefURL}
\newblock
{\href{https://arxiv.org/abs/https://doi.org/10.1287/moor.9.3.441}{{https://doi.org/10.1287/moor.9.3.441}}}
\newblock

\PrintBackRefs{\CurrentBib}

\bibitem [\protect \citeauthoryear {%
Sherzer%
}{%
Sherzer%
}{%
{\protect \APACyear {2025}}%
}]{%
SHERZER2025141}
\APACinsertmetastar {%
SHERZER2025141}%
\begin{APACrefauthors}%
Sherzer, E.%
\end{APACrefauthors}%
\unskip\
\newblock
\APACrefYearMonthDay{2025}{}{}.
\newblock
{\BBOQ}\APACrefatitle {Computing the steady-state probabilities of the number of customers in the system of a tandem queueing system, a Machine Learning approach} {Computing the steady-state probabilities of the number of customers in the system of a tandem queueing system, a machine learning approach}.{\BBCQ}
\newblock
\APACjournalVolNumPages{European Journal of Operational Research}{326}{1}{141-156,}
\newblock
\begin{APACrefDOI} \doi{https://doi.org/10.1016/j.ejor.2025.04.040} \end{APACrefDOI}
\newblock
\begin{APACrefURL} {https://www.sciencedirect.com/science/article/pii/S037722172500325X} \end{APACrefURL}
\newblock

\newblock

\PrintBackRefs{\CurrentBib}

\bibitem [\protect \citeauthoryear {%
Sherzer%
\ \protect \BOthers {.}}{%
Sherzer%
\ \protect \BOthers {.}}{%
{\protect \APACyear {2025}}%
}]{%
SHERZER2025889}
\APACinsertmetastar {%
SHERZER2025889}%
\begin{APACrefauthors}%
Sherzer, E.%
, Baron, O.%
, Krass, D.%
\BCBL {} Resheff, Y.%
\end{APACrefauthors}%
\unskip\
\newblock
\APACrefYearMonthDay{2025}{}{}.
\newblock
{\BBOQ}\APACrefatitle {Approximating G(t)/GI/1 queues with deep learning} {Approximating g(t)/gi/1 queues with deep learning}.{\BBCQ}
\newblock
\APACjournalVolNumPages{European Journal of Operational Research}{322}{3}{889-907,}
\newblock
\begin{APACrefDOI} \doi{https://doi.org/10.1016/j.ejor.2024.12.030} \end{APACrefDOI}
\newblock
\begin{APACrefURL} {https://www.sciencedirect.com/science/article/pii/S037722172400972X} \end{APACrefURL}
\newblock

\newblock

\PrintBackRefs{\CurrentBib}

\bibitem [\protect \citeauthoryear {%
Tan%
\ \BBA {} Khayyati%
}{%
Tan%
\ \BBA {} Khayyati%
}{%
{\protect \APACyear {2022}}%
}]{%
doi:10.1080/00207543.2021.1887536}
\APACinsertmetastar {%
doi:10.1080/00207543.2021.1887536}%
\begin{APACrefauthors}%
Tan, B.%
\BCBT {}\ \BBA {} Khayyati, S.%
\end{APACrefauthors}%
\unskip\
\newblock
\APACrefYearMonthDay{2022}{}{}.
\newblock
{\BBOQ}\APACrefatitle {Supervised learning-based approximation method for single-server open queueing networks with correlated interarrival and service times} {Supervised learning-based approximation method for single-server open queueing networks with correlated interarrival and service times}.{\BBCQ}
\newblock
\APACjournalVolNumPages{International Journal of Production Research}{60}{22}{6822--6847,}
\newblock
\begin{APACrefDOI} \doi{10.1080/00207543.2021.1887536} \end{APACrefDOI}
\newblock
\begin{APACrefURL} {https://doi.org/10.1080/00207543.2021.1887536} \end{APACrefURL}
\newblock
{\href{https://arxiv.org/abs/https://doi.org/10.1080/00207543.2021.1887536}{{https://doi.org/10.1080/00207543.2021.1887536}}}
\newblock

\PrintBackRefs{\CurrentBib}

\bibitem [\protect \citeauthoryear {%
Torab%
\ \BBA {} Kamen%
}{%
Torab%
\ \BBA {} Kamen%
}{%
{\protect \APACyear {2001}}%
}]{%
Torab2001}
\APACinsertmetastar {%
Torab2001}%
\begin{APACrefauthors}%
Torab, P.%
\BCBT {}\ \BBA {} Kamen, E.W.%
\end{APACrefauthors}%
\unskip\
\newblock
\APACrefYearMonthDay{2001}{}{}.
\newblock
{\BBOQ}\APACrefatitle {On Approximate Renewal Models for the Superposition of Renewal Processes} {On approximate renewal models for the superposition of renewal processes}.{\BBCQ}
\newblock
\APACjournalVolNumPages{IEEE Transactions on Automatic Control}{46}{6}{895--908,}
\newblock

\newblock

\PrintBackRefs{\CurrentBib}

\bibitem [\protect \citeauthoryear {%
Wagner%
}{%
Wagner%
}{%
{\protect \APACyear {1994}}%
}]{%
Wagner1994}
\APACinsertmetastar {%
Wagner1994}%
\begin{APACrefauthors}%
Wagner, D.%
\end{APACrefauthors}%
\unskip\
\newblock
\APACrefYearMonthDay{1994}{}{}.
\newblock
{\BBOQ}\APACrefatitle {Analysis of a Multi-Server Model with Non-Preemptive Priorities and Non-Renewal Input} {Analysis of a multi-server model with non-preemptive priorities and non-renewal input}.{\BBCQ}
\newblock
 \APACrefbtitle {Queueing and Related Models} {Queueing and related models}\ (\BPGS\ 457--480).
\newblock
\APACaddressPublisher{}{North-Holland}.
\PrintBackRefs{\CurrentBib}

\bibitem [\protect \citeauthoryear {%
Whitt%
}{%
Whitt%
}{%
{\protect \APACyear {1982}}%
}]{%
Whitt1982}
\APACinsertmetastar {%
Whitt1982}%
\begin{APACrefauthors}%
Whitt, W.%
\end{APACrefauthors}%
\unskip\
\newblock
\APACrefYearMonthDay{1982}{}{}.
\newblock
{\BBOQ}\APACrefatitle {Approximating a Point Process by a Renewal Process: Two Basic Methods} {Approximating a point process by a renewal process: Two basic methods}.{\BBCQ}
\newblock
\APACjournalVolNumPages{Operations Research}{30}{1}{125--147,}
\newblock

\newblock

\PrintBackRefs{\CurrentBib}

\bibitem [\protect \citeauthoryear {%
Whitt%
\ \BBA {} You%
}{%
Whitt%
\ \BBA {} You%
}{%
{\protect \APACyear {2019}}%
}]{%
WHITT201999}
\APACinsertmetastar {%
WHITT201999}%
\begin{APACrefauthors}%
Whitt, W.%
\BCBT {}\ \BBA {} You, W.%
\end{APACrefauthors}%
\unskip\
\newblock
\APACrefYearMonthDay{2019}{}{}.
\newblock
{\BBOQ}\APACrefatitle {The advantage of indices of dispersion in queueing approximations} {The advantage of indices of dispersion in queueing approximations}.{\BBCQ}
\newblock
\APACjournalVolNumPages{Operations Research Letters}{47}{2}{99-104,}
\newblock
\begin{APACrefDOI} \doi{https://doi.org/10.1016/j.orl.2019.01.001} \end{APACrefDOI}
\newblock
\begin{APACrefURL} {https://www.sciencedirect.com/science/article/pii/S016763771830333X} \end{APACrefURL}
\newblock

\newblock

\PrintBackRefs{\CurrentBib}

\bibitem [\protect \citeauthoryear {%
Whitt%
\ \BBA {} You%
}{%
Whitt%
\ \BBA {} You%
}{%
{\protect \APACyear {2022}}%
}]{%
https://doi.org/10.1002/nav.22010}
\APACinsertmetastar {%
https://doi.org/10.1002/nav.22010}%
\begin{APACrefauthors}%
Whitt, W.%
\BCBT {}\ \BBA {} You, W.%
\end{APACrefauthors}%
\unskip\
\newblock
\APACrefYearMonthDay{2022}{}{}.
\newblock
{\BBOQ}\APACrefatitle {A robust queueing network analyzer based on indices of dispersion} {A robust queueing network analyzer based on indices of dispersion}.{\BBCQ}
\newblock
\APACjournalVolNumPages{Naval Research Logistics (NRL)}{69}{1}{36-56,}
\newblock
\begin{APACrefDOI} \doi{https://doi.org/10.1002/nav.22010} \end{APACrefDOI}
\newblock
\begin{APACrefURL} {https://onlinelibrary.wiley.com/doi/abs/10.1002/nav.22010} \end{APACrefURL}
\newblock
{\href{https://arxiv.org/abs/https://onlinelibrary.wiley.com/doi/pdf/10.1002/nav.22010}{{https://onlinelibrary.wiley.com/doi/pdf/10.1002/nav.22010}}}
\newblock

\PrintBackRefs{\CurrentBib}

\bibitem [\protect \citeauthoryear {%
Zhang%
}{%
Zhang%
}{%
{\protect \APACyear {2018}}%
}]{%
8624183}
\APACinsertmetastar {%
8624183}%
\begin{APACrefauthors}%
Zhang, Z.%
\end{APACrefauthors}%
\unskip\
\newblock
\APACrefYearMonthDay{2018}{}{}.
\newblock
{\BBOQ}\APACrefatitle {Improved Adam Optimizer for Deep Neural Networks} {Improved adam optimizer for deep neural networks}.{\BBCQ}
\newblock
 \APACrefbtitle {2018 IEEE/ACM 26th International Symposium on Quality of Service (IWQoS)} {2018 ieee/acm 26th international symposium on quality of service (iwqos)}\ (\BPG~1-2).
\PrintBackRefs{\CurrentBib}

\end{thebibliography}

\begin{appendices}

\section{Generating MAPs}\label{append:gen_map}

In this section, we provide the MAP generating Algorithms with respect to Section~\ref{sec:data_generation}.

\begin{algorithm}[!ht]
\caption{\textsc{MAP with Strong Negative Autocorrelation}}
\label{alg:map_strong_negative}
\begin{algorithmic}[1]
\Require Fast/slow means $\mu_F,\mu_S$, Erlang orders $k_F,k_S$
\Ensure MAP $(D_0,D_1)$ with $\rho_1$ close to $-1$ and $\mathbb{E}[A]=1$

\State \textbf{Construct regime PH distributions}
\State $(\alpha_F,T_F) \leftarrow \textsc{ErlangPH}(k_F,\ \text{mean}=\mu_F)$
\State $(\alpha_S,T_S) \leftarrow \textsc{ErlangPH}(k_S,\ \text{mean}=\mu_S)$

\State \textbf{Define regimes}
\State $\mathcal{R} \leftarrow \{(\alpha_F,T_F),(\alpha_S,T_S)\}$

\State \textbf{Deterministic alternation}
\State $R \leftarrow 
\begin{pmatrix}
0 & 1 \\
1 & 0
\end{pmatrix}$

\State \textbf{Build regime-switching MAP}
\State $(D_0,D_1) \leftarrow \textsc{BuildRegimeSwitchingMAP}(\mathcal{R},R)$

\State \textbf{Scale to unit mean}
\State $(D_0,D_1) \leftarrow \textsc{TimeScaleMAP}(D_0,D_1,\ \text{targetMean}=1)$

\State \Return $(D_0,D_1)$
\end{algorithmic}
\end{algorithm}

\begin{algorithm}[!ht]
\caption{\textsc{MAP with Strong Positive Autocorrelation}}
\label{alg:map_strong_positive}
\begin{algorithmic}[1]
\Require Fast/slow means $\mu_F,\mu_S$, Erlang orders $k_F,k_S$, stickiness $p_{\mathrm{stay}}$
\Ensure MAP $(D_0,D_1)$ with $\rho_1$ close to $+1$ and $\mathbb{E}[A]=1$

\State \textbf{Construct regime PH distributions}
\State $(\alpha_F,T_F) \leftarrow \textsc{ErlangPH}(k_F,\ \text{mean}=\mu_F)$
\State $(\alpha_S,T_S) \leftarrow \textsc{ErlangPH}(k_S,\ \text{mean}=\mu_S)$

\State \textbf{Define regimes}
\State $\mathcal{R} \leftarrow \{(\alpha_F,T_F),(\alpha_S,T_S)\}$

\State \textbf{Sticky regime transition matrix}
\State $R \leftarrow 
\begin{pmatrix}
p_{\mathrm{stay}} & 1-p_{\mathrm{stay}} \\
1-p_{\mathrm{stay}} & p_{\mathrm{stay}}
\end{pmatrix}$

\State \textbf{Build regime-switching MAP}
\State $(D_0,D_1) \leftarrow \textsc{BuildRegimeSwitchingMAP}(\mathcal{R},R)$

\State \textbf{Scale to unit mean}
\State $(D_0,D_1) \leftarrow \textsc{TimeScaleMAP}(D_0,D_1,\ \text{targetMean}=1)$

\State \Return $(D_0,D_1)$
\end{algorithmic}
\end{algorithm}

\begin{algorithm}[!ht]
\caption{\textsc{MAP with Moderate Correlation and Varied Marginals}}
\label{alg:map_medium_corr_varied_marginals}
\begin{algorithmic}[1]
\Require Target dependence $\rho_{\mathrm{target}}$, regime means $\mu_F=\texttt{mean\_fast}$, $\mu_S=\texttt{mean\_slow}$, flag $\texttt{heavy\_marginals}$, optional seed
\Ensure MAP matrices $(D_0,D_1)$ with $\mathbb{E}[A]=1$ and moderate $\rho_1$ (sign/magnitude guided by $\rho_{\mathrm{target}}$)

\State Initialize RNG with \texttt{seed}

\State \textbf{Fast regime PH (typically low variability):}
\State Sample $k_F \sim \mathrm{UnifInt}\{3,\dots,19\}$
\State $(\alpha_F, T_F) \leftarrow \textsc{ErlangPH}(k_F,\ \text{mean}=\mu_F)$

\State \textbf{Slow regime PH (heavier / more flexible marginals):}
\If{$\texttt{heavy\_marginals}=\texttt{True}$}
    \State Sample $p \sim \mathrm{Unif}(0.2,0.8)$ and $r \sim \mathrm{Unif}(5,80)$
    \State $(\alpha_H, T_H) \leftarrow \textsc{HyperExp2PH}(\text{mean}=\mu_S,\ \text{mix}=p,\ \text{rateRatio}=r)$
    \If{$\mathrm{Unif}(0,1) < 0.6$}
        \State Sample $k_S \sim \mathrm{UnifInt}\{2,\dots,14\}$ and $w \sim \mathrm{Unif}(0.2,0.8)$
        \State $(\alpha_E, T_E) \leftarrow \textsc{ErlangPH}(k_S,\ \text{mean}=\mu_S)$
        \State $(\alpha_S, T_S) \leftarrow \textsc{MixPH}((\alpha_H,T_H),(\alpha_E,T_E),\ w)$
    \Else
        \State $(\alpha_S, T_S) \leftarrow (\alpha_H, T_H)$
    \EndIf
\Else
    \State Sample $k_S \sim \mathrm{UnifInt}\{2,\dots,24\}$
    \State $(\alpha_S, T_S) \leftarrow \textsc{ErlangPH}(k_S,\ \text{mean}=\mu_S)$
\EndIf

\State \textbf{Regimes:} $\mathcal{R} \leftarrow \{(\alpha_F,T_F),\ (\alpha_S,T_S)\}$

\State \textbf{Map $\rho_{\mathrm{target}}$ to regime stickiness:}
\State $p_{\mathrm{stay}} \leftarrow 0.5 + 0.45 \tanh(2\rho_{\mathrm{target}})$
\State $p_{\mathrm{stay}} \leftarrow \min\{0.95,\max\{0.05,\ p_{\mathrm{stay}}\}\}$

\State \textbf{Regime transition matrix:}
\State $R \leftarrow \begin{pmatrix}
p_{\mathrm{stay}} & 1-p_{\mathrm{stay}}\\
1-p_{\mathrm{stay}} & p_{\mathrm{stay}}
\end{pmatrix}$

\State \textbf{Build MAP from regime switching:}
\State $(D_0,D_1) \leftarrow \textsc{BuildRegimeSwitchingMAP}(\mathcal{R}, R)$

\State \textbf{Scale time to unit mean:}
\State $(D_0,D_1) \leftarrow \textsc{TimeScaleMAP}(D_0,D_1,\ \text{targetMean}=1)$

\State \Return $(D_0,D_1)$
\end{algorithmic}
\end{algorithm}

\section{Hyper-param finetune}\label{append:finetune}

We present the technical details of the hyperparameter tuning procedure for a fixed pair $(n_{\text{arrival}}, n_{\text{service}})$. The hyperparameters optimized in our study include the learning rate, number of training epochs, number of hidden layers, number of neurons per layer, batch size, and the weight decay parameter of the Adam optimizer. The latter serves as a regularization term; see~\cite{8624183} and the references therein for additional details on the Adam optimization method.

The search domain for each hyperparameter is defined as follows:
\begin{itemize}
    \item Learning rate: $\{0.01, 0.005, 0.001, 0.0001\}$.
    \item Number of training epochs: $[100, 150]$.
    \item Number of hidden layers: $[3, 6]$.
    \item Number of neurons per layer: $\{50, 60, 70\}$.
    \item Batch size: $\{64, 128, 256\}$.
    \item Weight decay: $\{10^{-4}, 10^{-5}, 10^{-6}\}$.
\end{itemize}

We perform 500 random search iterations over this hyperparameter space and select the configuration that achieves the lowest SAE on the validation set.

Each search requires approximately 0.76 hours on average, using an NVIDIA GeForce RTX 4070 Tensor Core GPU with 8GB of memory.

\section{Trained NN model properties}\label{append:architecture}

The superposition NN model contains 5 hidden layers, with sizes: (50,50,70,70,50),  totaling 11224 parameters. The selected batch size was 64, the weight decay value is $10^{-5}$,  learning rate $0.0001$, trained over 150 epochs.  

\end{appendices}


\end{document}